\documentclass[journal]{IEEEtran}
\usepackage{cite}

\ifCLASSINFOpdf
  \usepackage[pdftex]{graphicx}
  \graphicspath{{./graphic/}}
  \DeclareGraphicsExtensions{.pdf,.jpeg,.png,.jpg}
\else
\fi
\usepackage{amsmath}
\usepackage{url}

\usepackage{siunitx} 
\usepackage[dvipsnames,table,xcdraw]{xcolor} 
\usepackage{pifont} 
\usepackage{tensor} 
\usepackage{arydshln} 
\usepackage{amssymb} 

\newcommand{\abs}[1]{\left\lvert#1\right\rvert} 

\DeclareMathOperator{\mean}{mean}
\DeclareMathOperator{\std}{std}
\DeclareMathOperator{\med}{med}

\DeclareMathOperator*{\argmax}{arg\,max}
\DeclareMathOperator*{\argmin}{arg\,min}
\DeclareMathOperator{\atantwo}{atan2} 

\begin{document}
%
\title{Best Viewpoints for External Robots or Sensors Assisting Other Robots}
%
%
%

\author{Jan~Dufek,
        Xuesu~Xiao,
        and~Robin~R.~Murphy,~\IEEEmembership{Fellow,~IEEE}
\thanks{J. Dufek and R. R. Murphy are with the Department
of Computer Science and Engineering, Texas A\&M University, College Station,
TX, 77843 USA e-mail: dufek@tamu.edu and murphy@cse.tamu.edu.}
\thanks{X. Xiao is with the Department
of Computer Science, The University of Texas at Austin, Austin,
TX, 78712 USA e-mail: xiao@cs.utexas.edu.}
\thanks{This work was supported by NSF 1945105 Best Viewpoints for External Robots or Sensors Assisting Other Robots and DOE {DE-EM0004483} NRI: A Collaborative Visual Assistant for Robot Operations in Unstructured or Confined Environments.}
\thanks{\copyright~2021 IEEE. Personal use of this material is permitted. Permission from IEEE must be obtained for all other uses, in any current or future media, including reprinting/republishing this material for advertising or promotional purposes, creating new collective works, for resale or redistribution to servers or lists, or reuse of any copyrighted component of this work in other works.}}

%
%

\markboth{IEEE Transactions on Human-Machine Systems}%
{Dufek \MakeLowercase{\textit{et al.}}: Best Viewpoints for External Robots or Sensors Assisting Other Robots}
%



\maketitle

\begin{abstract}
This work creates a model of the value of different external viewpoints of a robot performing tasks. The current state of the practice is to use a teleoperated assistant robot to provide a view of a task being performed by a primary robot; however, the choice of viewpoints is ad hoc and does not always lead to improved performance. This research applies a psychomotor approach to develop a model of the relative quality of external viewpoints using Gibsonian affordances. In this approach, viewpoints for the affordances are rated based on the psychomotor behavior of human operators and clustered into manifolds of viewpoints with the equivalent value. The value of 30 viewpoints is quantified in a study with 31 expert robot operators for 4 affordances (\textsc{Reachability}, \textsc{Passability}, \textsc{Manipulability}, and \textsc{Traversability}) using a computer-based simulator of two robots. The adjacent viewpoints with similar values are clustered into ranked manifolds using agglomerative hierarchical clustering. The results show the validity of the affordance-based approach by confirming that there are manifolds of statistically significantly different viewpoint values, viewpoint values are statistically significantly dependent on the affordances, and viewpoint values are independent of a robot. Furthermore, the best manifold for each affordance provides a statistically significant improvement with a large Cohen's $d$ effect size (1.1--2.3) in performance (improving time by 14\%--59\% and reducing errors by 87\%--100\%) and improvement in performance variation over the worst manifold. This model will enable autonomous selection of the best possible viewpoint and path planning for the assistant robot.
\end{abstract}


%
\IEEEpeerreviewmaketitle

\section{Introduction}
\label{sec:introduction}

\IEEEPARstart{A}{n} assistant robot providing a view of a task being performed by a primary robot has emerged as the state of the practice for ground and water robots in homeland security applications, disaster response, and inspection tasks~\cite{murphy2019introduction,dufek2016visual,xiao2017uav,dufek2017visual,dufek2018theoretical,dufek2019visual}. Advances in small unmanned aerial systems (UAS), especially tethered UAS, suggest that flying assistant robots will soon supply the needed external visual perspective~\cite{xiao2017visual,xiao2018indoor,xiao2019benchmarking,xiao2018motion}.

During the 2011 Fukushima Daiichi nuclear power plant accident, teleoperated robots were used in pairs from the beginning of the response to reduce the time it took to accomplish a task~\cite{guizzo2011fukushima,murphy2014disaster}. iRobot PackBot unmanned ground vehicles (UGVs) were used to conduct radiation surveys and read dials inside the plant facility, where the assistant PackBot provided camera views of the first robot in order to manipulate door handles, valves, and sensors faster~\cite{kawatsuma2012emergency}.

Since then, the use of two robots to perform a single task has been acknowledged as the best practice for decommissioning tasks. However, the Japanese Atomic Energy Agency (JAEA) has reported through our memorandum of understanding for cooperative research on disaster robotics that operators constantly try to avoid using a robotic visual assistant. The two sets of robot operators find it difficult to coordinate in order to get and maintain the desired view but a single operator becomes frustrated trying to operate both robots.

There are at least two issues with the current state of the practice. First, it increases the cognitive workload on a primary operator by either requiring the primary operator to control two robots or having to coordinate with a secondary operator~\cite{murphy2014disaster}. Second, it is not guaranteed a human operator will provide ideal viewpoints as viewpoint quality for various tasks is not well understood and humans were shown to pick suboptimal viewpoints~\cite{mckee2003human}.

This article addresses the choice of ideal viewpoints by creating a model of the value of different external viewpoints of a robot performing tasks; it is expected, but beyond the scope of this study, that the application of the model to robotic visual assistants will likely reduce the cognitive workload on the primary operator. The model will provide an understanding of the utility of different external viewpoints of tasks of the primary robot and can be used as a basis for principled viewpoint selection for a robotic visual assistant. This can ultimately enable autonomous viewpoint selection and path planning for an autonomous robotic visual assistant, therefore, eliminating the need for manual control.

This article is organized as follows. Section~\ref{sec:related_work} discusses the related work establishing there is no existing model of viewpoint values and showing the importance of psychomotor aspects in viewpoint selection. Section~\ref{sec:approach} introduces the affordance-based approach. Section~\ref{sec:implementation} details the implementation of a computer-based simulator. Section~\ref{sec:experimentation} presents a human subject study quantifying the value of viewpoints and clustering to create the manifolds. Section~\ref{sec:results} presents the results showing the validity of the affordance-based approach and a significant improvement in performance. Section~\ref{sec:discussion} discusses the relation to the related work, the reduction in cognitive workload, the ramifications for robotic visual assistants, and the actionable rules for teleoperated robotic visual assistants. Section~\ref{sec:summary} summarizes the key findings that there are manifolds of different viewpoint values, viewpoint values are dependent on the affordances, and viewpoint values are independent of the robot.

\section{Related Work}
\label{sec:related_work}

There is no existing model of viewpoint values leaving robotic visual assistants to rely on ad hoc choices of viewpoints or work envelope models. Woods et al.~\cite{roesler2005a,morison2009integrating,morison2010perspective,morison2016seeing,murphy2013apprehending,murphy2016affordances,murphy2016can,murphy2017within,morison2015human,woods2004envisioning} indicated that improving the ability to comprehend Gibsonian affordances improves teleoperation and external viewpoints improve the ability to comprehend affordances forming an important foundation for the affordance-based approach of this work.

A total of five attributes of an ideal viewpoint were identified and four categories of existing robotic visual assistant implementations were examined with an underlying focus on whether there is an existing model of viewpoint values. There was no existing model of viewpoint values leaving the existing robotic visual assistant implementations to rely on ad hoc choices or on having a priori access to, or constructing, 2D or 3D models of the work envelope. Robotic visual assistants lacked principles to select ideal viewpoints and no robotic visual assistant implementation considered psychomotor aspects in the viewpoint selection.

There were five attributes of an ideal external viewpoint of action being performed by a robot: the field of view (the area of interest must be in the field of view)~\cite{gatesichapakorn2019ros}, visibility/occlusions (the view of the area of interest must be occlusion free)~\cite{yang2018viewpoint}, depth of field (the area of interest must be in the depth of field or sharp focus)~\cite{bonaventura2018a}, resolution/zoom (the area of interest must have sufficient resolution in the image so the camera has to be physically close or have to zoom in)~\cite{kurtser2018the}, and psychomotor aspects (the view must positively affect the human ability to move the robot to accomplish the goal)~\cite{murphy2017within}.

There were four categories of robotic visual assistant implementations in the literature all lacking principles to select ideal viewpoints. Static visual assistants~\cite{sato2019derivation,dima2019view,sato2019experimental} did not move and therefore could not adapt viewpoints to changing pose or actions of the primary robot. Manual visual assistants~\cite{leon2016from,kiribayashi2018design} left the choice of a viewpoint to humans who were previously shown to pick suboptimal viewpoints~\cite{mckee2003human}. Reactive autonomous visual assistants~\cite{nicolis2018occlusion,gawel2018aerial,chikushi2020automated} only reactively tracked and zoomed on the action ignoring the question of what are the best viewpoints. Deliberative autonomous visual assistants~\cite{samejima2018visual,thomason2019a,rakita2019remote} deliberated about certain predefined geometrical criteria while only considering camera configuration attributes of an ideal viewpoint (field of view, visibility/occlusions, depth of field, and resolution/zoom). While camera configuration attributes are necessary preconditions for an ideal viewpoint, no robotic visual assistant studies considered how the viewpoint affects the human teleoperator of the primary robot (psychomotor aspects attribute of an ideal viewpoint) in viewpoint selection.

Psychomotor aspects of an ideal viewpoint were ignored in the existing robotic visual assistant implementations, despite the results by Woods et al.~\cite{roesler2005a,morison2009integrating,morison2010perspective,morison2016seeing,murphy2013apprehending,murphy2016affordances,murphy2016can,murphy2017within,morison2015human,woods2004envisioning} who showed that teleoperation can be improved by improving the ability to comprehend affordances and that an external view improves the ability to comprehend affordances.

Woods et al.\ primarily focused on creating tools to enable humans to manually select external views that supply Gibsonian affordances, which are visual cues that allow humans to directly perceive the possibility of actions independent of the environment or task models~\cite{gibson2014ecological}. Woods et al.\ contributed two important results forming a theoretical background for this article. They showed that teleoperation can be improved by improving teleoperators' ability to comprehend affordances and they established that an external view improves the ability to comprehend the affordances. This indicates that the value of a viewpoint should depend on the affordances and confirms the benefit of a robotic visual assistant providing an external view.

Despite those contributions, Woods et al.\ relied on human input to select viewpoints and did not evaluate the value of different external viewpoints, they experimentally studied only \textsc{Reachability} affordance, they used a simulator that did not reflect realistic robots, and the subjects were not expert robot operators. Their work forms the foundation for the approach in Section~\ref{sec:approach}. However, unlike their work, this article creates a model of the value of different external viewpoints (that can ultimately enable a robotic visual assistant to pick a viewpoint without human input), the model is created for four affordances, the simulator used in the experimentation reflects two realistic robots, and the subjects are expert robot operators. Having expert operators using realistic robots prevents confounding the results with subjects struggling to control the robots.

\section{Affordance-Based Approach}
\label{sec:approach}

\begin{figure}[!t]
\centering
\includegraphics[width=0.49\textwidth]{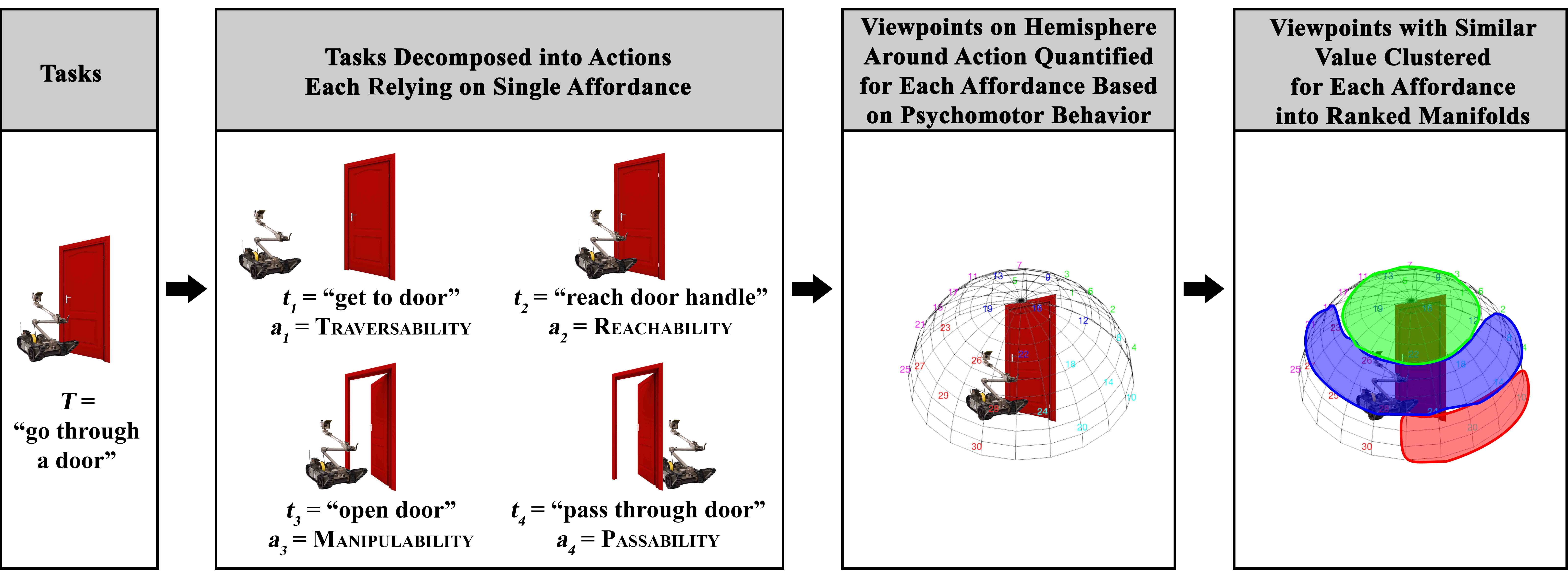}
\caption{An overview of the main building blocks of the approach. $T$, $t_i$, and $a_i$ denote task, action, and affordance respectively.}
\label{fig:approach}
\end{figure}

The approach is to use the concept of Gibsonian affordances~\cite{murphy2019introduction}, where the potential for an action can be directly perceived without knowing intent or models, and thus is universal to all robots and tasks. In this approach, it is assumed tasks can be decomposed into actions each relying on a single affordance, space around the actions is decomposed into viewpoints, and the viewpoints for the affordances are rated based on teleoperator's psychomotor behavior and clustered into manifolds of viewpoints with the equivalent value (Figure~\ref{fig:approach}).

The main postulation of the approach is that a model of viewpoint values can be created using Gibsonian affordances based on psychomotor behavior. This postulation has two central tenets. The first tenet is that the value of a viewpoint depends on the Gibsonian affordance for each action in a task. This tenet is supported by the previous work of Woods et al.\ discussed in Section~\ref{sec:related_work}. The approach based on Gibsonian affordances has at least two benefits. First, it avoids the need for models seen in the deliberative approach by focusing on the affordances for an action rather than the action itself. Second, research on affordances suggests there are relatively few affordances~\cite{hallford1984sizing}. It is therefore conceivable every robotic task could be decomposed into a small set of affordances and each affordance would have associated preferred viewpoints. The second tenet is that viewpoints in the space surrounding the action can be rated and adjacent viewpoints with similar ratings can be clustered into manifolds of viewpoints with the equivalent value. The clustering of viewpoints into manifolds has at least three benefits. First, it simplifies navigational reachability. As long as the robotic visual assistant can reach any location within the manifold, it will provide approximately the same value as any other location within the manifold. Second, it aids visual stability. Due to equivalence of viewpoints within the manifold, positioning the robotic visual assistant at the centroid of a manifold will minimize the chance that a potential pose perturbation would significantly change viewpoint quality. Third, it can be used in autonomous planning for a robotic visual assistant to select a manifold and plan a path there while balancing the reward of having a view from that particular manifold with the associated risk of being at that manifold and getting to that manifold~\cite{xiao2019explicit,xiao2019robot,xiao2019autonomous,xiao2019explicit2,xiao2020tethered,xiao2020robot} while also considering visual stability.

\begin{figure}[!t]
\centering
\includegraphics[width=2.5in]{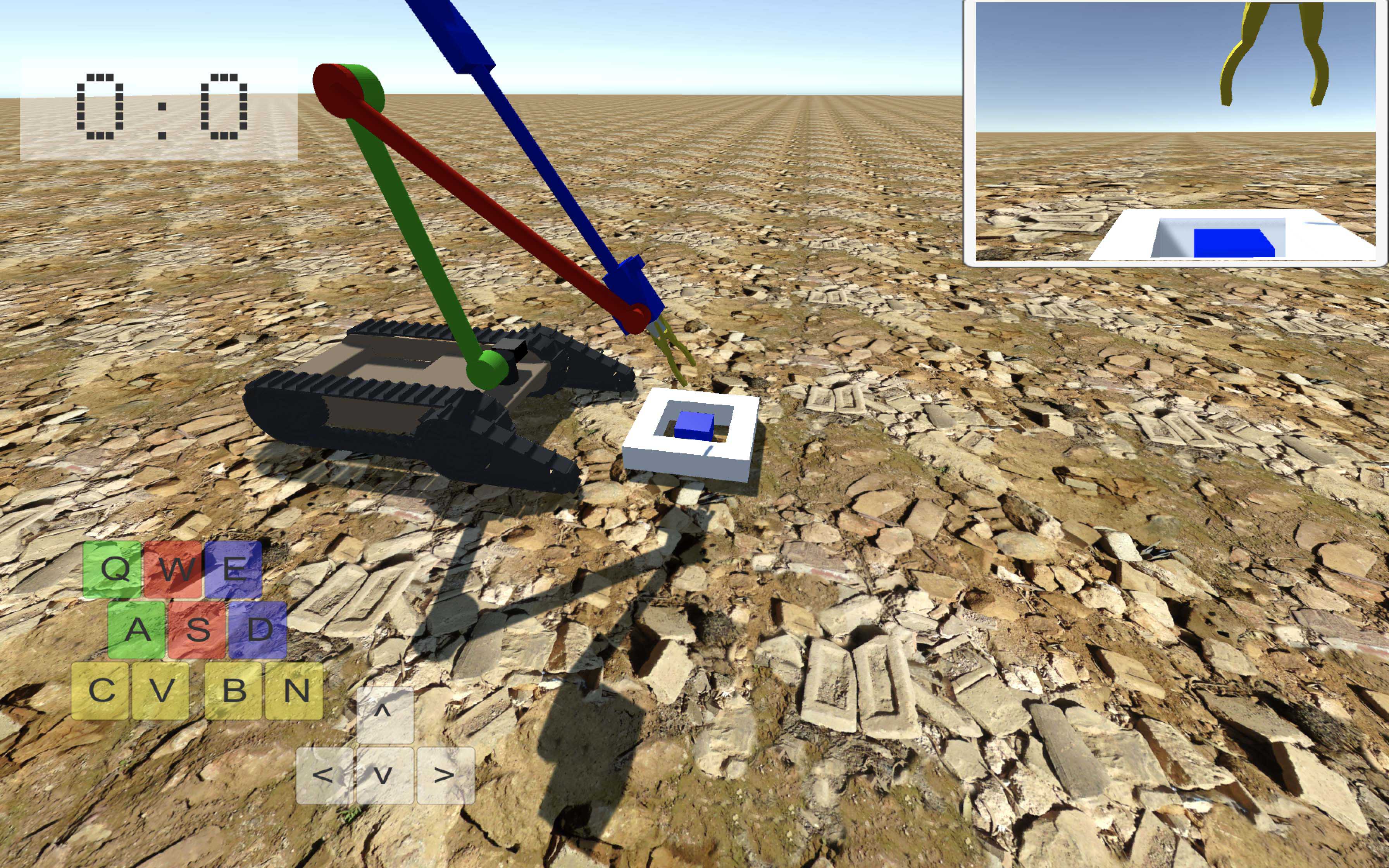}
\caption{\textsc{Reachability} affordance: Are the robot and its manipulator in the right pose to reach an object?}
\label{fig:reachability}
\end{figure}

\begin{figure}[!t]
\centering
\includegraphics[width=2.5in]{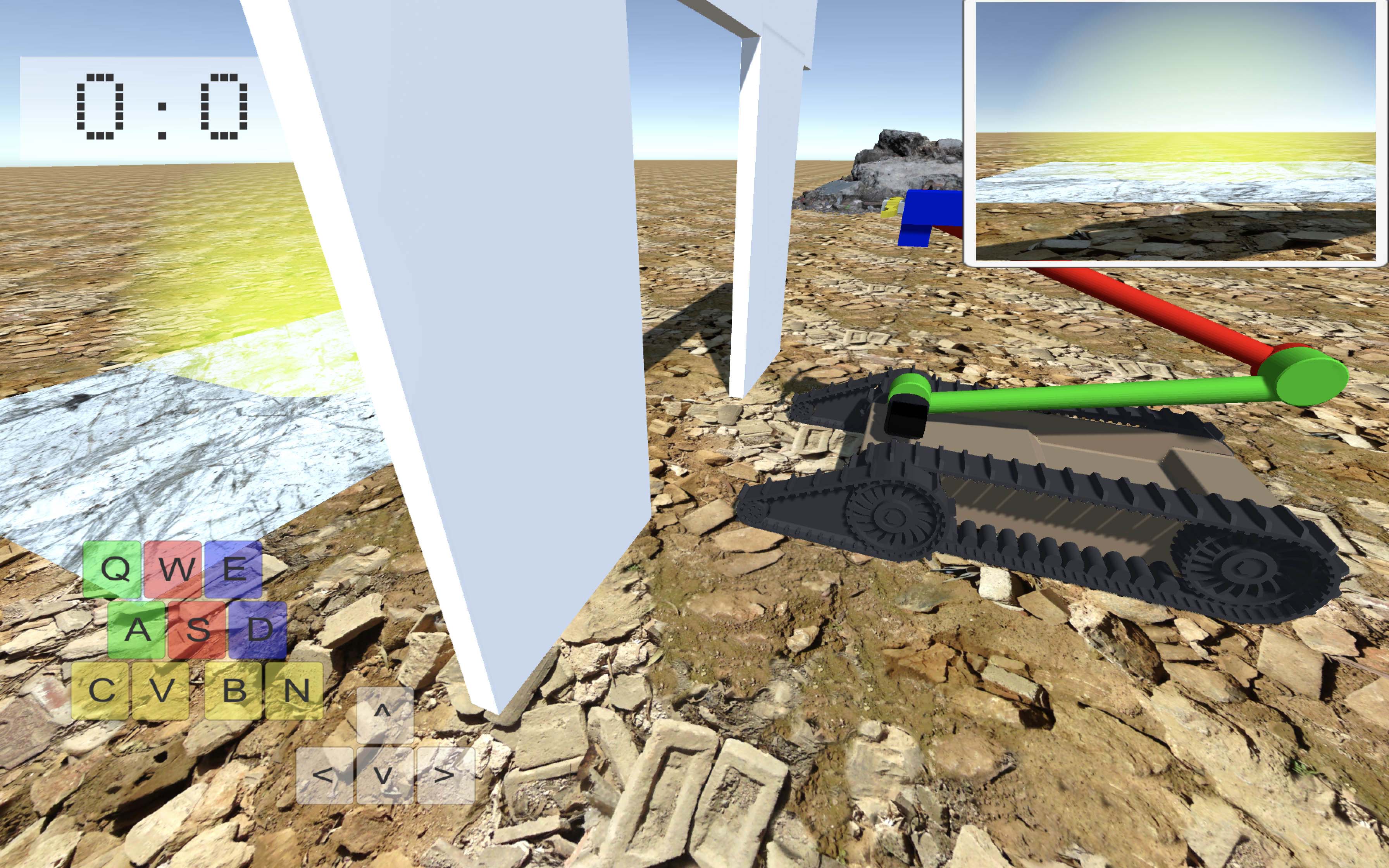}
\caption{\textsc{Passability} affordance: Is the robot or its manipulator in the right pose to safely pass through a narrow opening?}
\label{fig:passability}
\end{figure}

\begin{figure}[!t]
\centering
\includegraphics[width=2.5in]{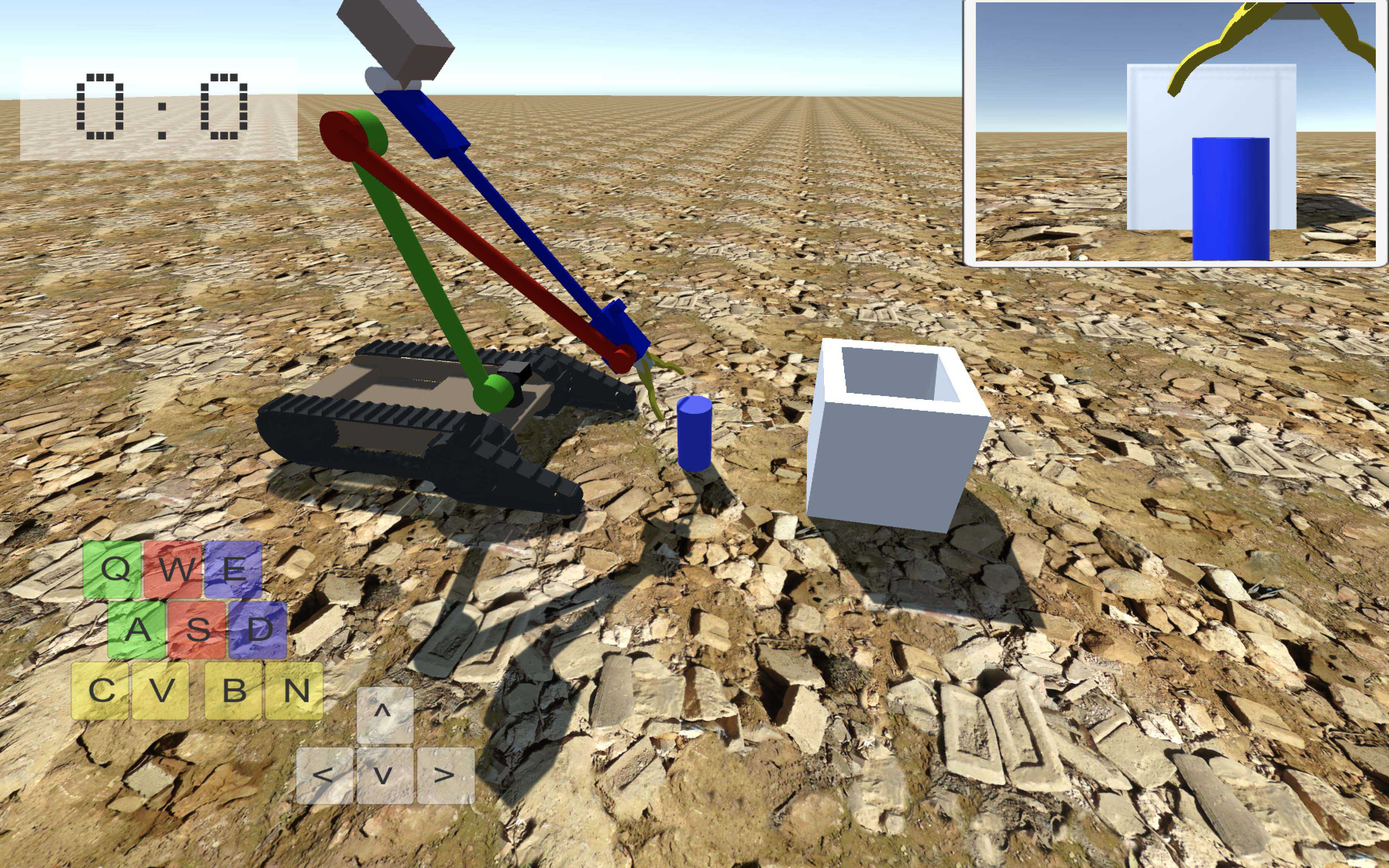}
\caption{\textsc{Manipulability} affordance: Is the robot's manipulator in the right pose to manipulate an object?}
\label{fig:manipulability}
\end{figure}

\begin{figure}[!t]
\centering
\includegraphics[width=2.5in]{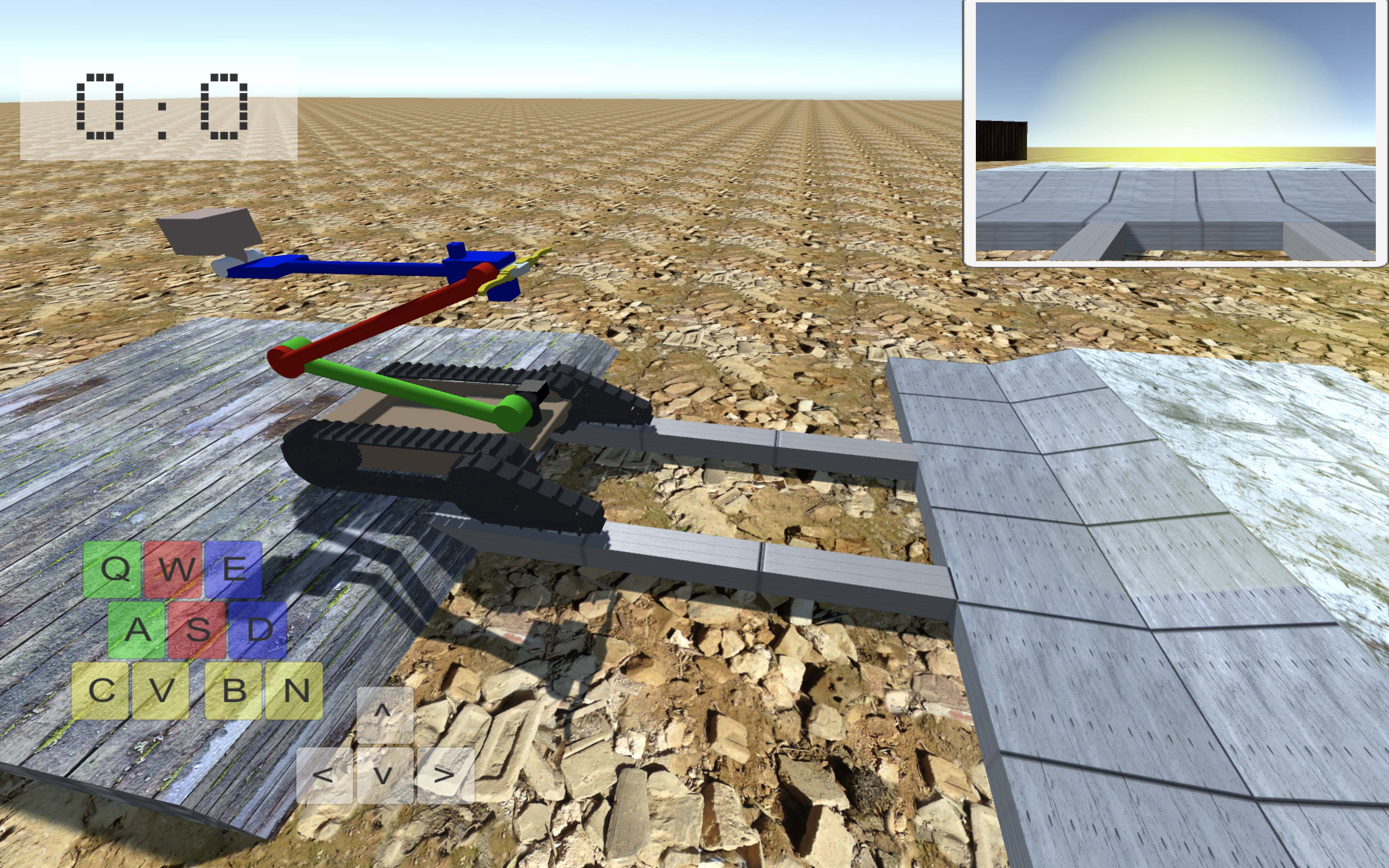}
\caption{\textsc{Traversability} affordance: Is the robot in the right pose to safely traverse the environment?}
\label{fig:traversability}
\end{figure}

Based on the related work of Woods et al.\ and our prior experience with 21 disaster deployments, participation in 35 homeland security exercises, and examination of common tasks for robots at Fukushima~\cite{murphy2014disaster}, the development of the model is restricted to four common affordances: \textsc{Reachability} (Figure~\ref{fig:reachability}), \textsc{Passability} (Figure~\ref{fig:passability}), \textsc{Manipulability} (Figure~\ref{fig:manipulability}), and \textsc{Traversability} (Figure~\ref{fig:traversability}). Woods et al.\ additionally discussed \textsc{Climability} and \textsc{Drivability} affordances, however, those overlap with our \textsc{Traversability} (definitions of affordances are not standardized).

Starting with the first building block from Figure~\ref{fig:approach}, it is assumed every task $T$ can be decomposed into a sequence of actions $t_1, t_2, ..., t_n$ where the perception for each action $t_i$ relies on a single affordance, $a_i$. In reality, actions might rely on a compound affordance, but this work assumes each action relies on its dominant affordance. Then a task $T$ can be represented by a sequence of action-affordance tuples $(t_i, a_i)$ forming a coarse knowledge representation of the task.

\begin{figure}[!t]
\centering
\includegraphics[width=2.5in]{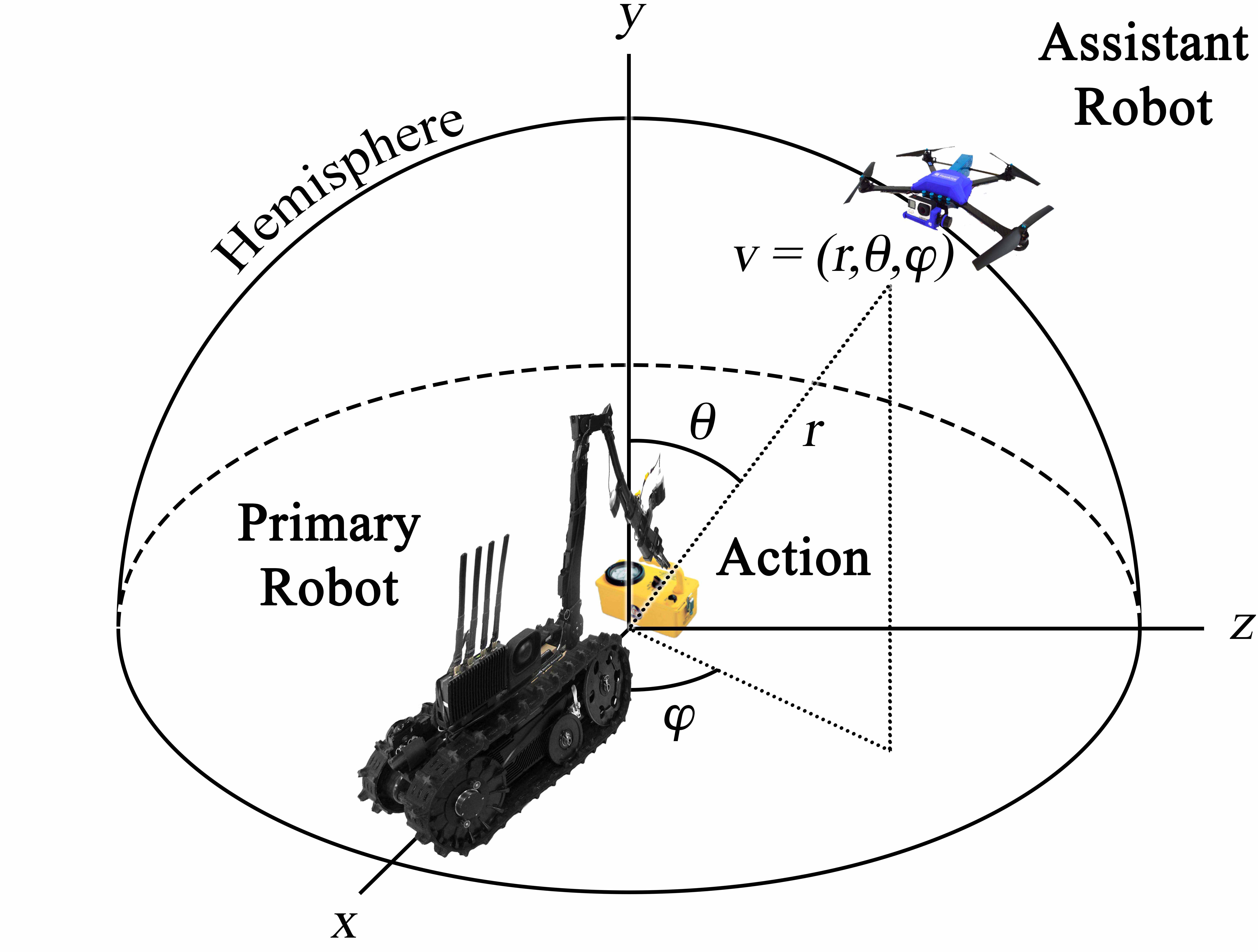}
\caption{A hemisphere with a fixed radius $r$ centered in the action serves as an idealized work envelope for the robotic visual assistant.}
\label{fig:hemisphere}
\end{figure}

Space around the action can be decomposed into viewpoints that are assumed to be lying on a hemisphere of a fixed radius (Figure~\ref{fig:hemisphere}). A viewpoint is represented using a spherical coordinate system as $v = (r, \theta, \varphi)$ and the optical axis is along the radius $r$. While the values of $r$ can vary in practice, an assumption for this work is that a hemisphere with a fixed radius $r$ serves as the idealized workspace envelope for the assistant.

A viewpoint $v$ will have a value $\abs{v}$ based on how well a teleoperator can perform the action from that viewpoint. The value is composed of the time to complete the action and the number of errors.

Adjacent viewpoints $v$ with similar value $\abs{v}$ will form a continuous volume, or manifold, $M$. Within a manifold, each viewpoint is equally good. The entire space will be divided into ranked manifolds.

The model of viewpoint values will be extracted in two steps. First, the value of viewpoints $\abs{v}$ for the four affordances will be quantified in a human subject study using a computer-based simulation. Second, adjacent viewpoints of similar value will be clustered into manifolds of viewpoints with the equivalent value.

\section{Simulator Implementation}
\label{sec:implementation}

A computer-based simulator was created to enable the quantification of the value of viewpoints by remotely (over the web) measuring the performance of expert robot operators controlling one of two robots (iRobot PackBot or QinetiQ TALON) in four tasks corresponding to the four affordances from different external viewpoints. Using expert robot operators already proficient with the robots reduces the chance of confounding the results with subjects’ varying familiarity with the robots, varying difficulty in controlling the robots, and varying time needed to train on the robots. It also reduces learning effects as subjects unfamiliar with the robots might gradually learn how to control the robots during the experiment. Those two specific models of robots were selected because they are the two most common explosive ordnance disposal robots making it easier to find subjects proficient with at least one of them. The use of computer-based simulation is justified based on previous work of Woods et al.\ (Section~\ref{sec:related_work}) who showed computer-based simulation is suitable to measure the teleoperators' ability to comprehend Gibsonian affordances. The simulator was implemented in C\# using the Unity engine and runs on Amazon Web Services (AWS) infrastructure. The AWS S3 supports a front-end website with the Unity simulation interface while AWS EC2 runs a back-end responsible for receiving and storing the data. When running the simulation, subjects can see a large external view of the task from a specific viewpoint, a small fixed view from a forward-looking onboard camera of the primary robot, a color-coded keyboard legend corresponding to the color-coding of the primary's robot arm (this is necessary because a keyboard is not a typical mode of control of those robots), and a clock to constantly remind them they are being timed (as seen in Figures~\ref{fig:reachability}--\ref{fig:traversability}). When a subject makes an error, the error location is highlighted in red and an error sound is played to make the subject aware of the error.

\section{Experimentation}
\label{sec:experimentation}

The experimentation is done by quantifying the value of viewpoints in a human subject study and then clustering the viewpoints into ranked manifolds. The value of 30 viewpoints is quantified in a 31 person human subject study for 4 Gibsonian affordances (\textsc{Reachability}, \textsc{Passability}, \textsc{Manipulability}, and \textsc{Traversability}) using a computer-based simulator. The data from the human subject study are then used to rate the viewpoints and cluster adjacent viewpoints with similar value into manifolds of viewpoints with the equivalent value using agglomerative hierarchical clustering.

\subsection{Quantifying Viewpoints in Human Subject Study}
\label{sec:experimentation_study}

A 31 person human subjects study was designed with a goal to sufficiently sample human performance for 30 viewpoints $v_i$, where $i = 1,...,30$, to quantify the value of viewpoints $\abs{v_i^a}$ for each of the 4 affordances $a$ so that spatial clusters (manifolds) can be learned. The subjects perform 4 tasks corresponding to the 4 affordances from varying external viewpoints while their performance is measured in terms of time and number of errors to quantify the corresponding viewpoint value.

The subjects are 31 (based on power analysis) male expert robot operators of age ranging from $23$ to $46$ years ($M = 31.5$, $SD = 5.9$) experienced with either PackBot or TALON robots. The subjects use their own computer to connect to a remote computer-based simulator via a web browser. The subjects choose either PackBot or TALON robots based on their experience (10 subjects chose PackBot and 21 chose TALON).

The subjects perform four kinds of tasks each associated with one of the four affordances. For \textsc{Reachability}, the task is to touch the blue cube using the gripper without hitting the neighboring blocks (Figure~\ref{fig:reachability}). For \textsc{Passability}, the task is to pass through the opening in the walls and take caution to not hit the walls (Figure~\ref{fig:passability}). For \textsc{Manipulability}, the task is to pick up the blue cylinder and drop it in the bin without hitting the bin with the gripper (Figure~\ref{fig:manipulability}). For \textsc{Traversability}, the task is to cross the ridge and reach the other side without falling on the ground (Figure~\ref{fig:traversability}).

\begin{figure}[!t]
\centering
\includegraphics[width=2.5in]{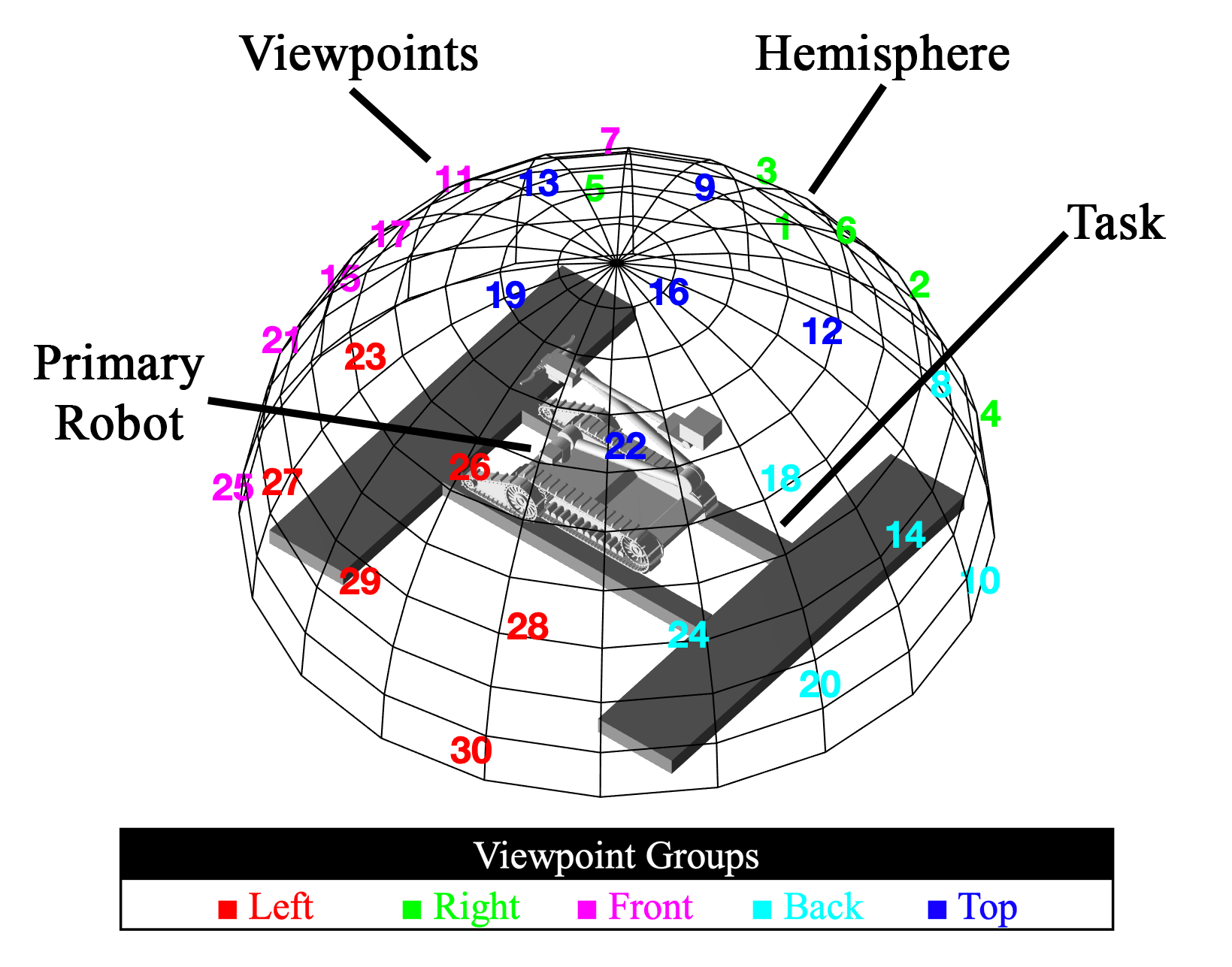}
\caption{A total of 30 possible viewpoints are equidistantly dispersed on a hemisphere centered at the task and divided into 5 groups. The figure is in scale.}
\label{fig:viewpoints}
\end{figure}

The independent variable is the position of the external viewpoint provided to the subject. A total of 30 possible viewpoints, $v_i$ where $i = 1,...,30$, are equidistantly dispersed on a hemisphere with a fixed radius of $r =$~\SI{1.5}{\meter} centered at the task location at $(0,0,0)$ as illustrated in Figure~\ref{fig:viewpoints}. The distance between viewpoints is approximately \SI{0.7}{\meter}. Those 30 viewpoints are divided into 5 groups (6 viewpoints per group) based on their relative position to the task location: left, right, front, back, and top. Each subject performs each of the 4 tasks from each of the 5 viewpoints groups (20 rounds total). The particular viewpoint from each group is always selected randomly. The order of the tasks and viewpoint groups is randomized for each subject to reduce the order effect. The 5 viewpoint groups are used solely to help the samples to be uniformly distributed across viewpoints and the specific choice of groups does not have an influence on the final results. An alternative approach would be to manually distribute the viewpoints to the subjects in a way that each viewpoint has the same number of samples.

There are two dependent variables both indicating the subject's performance: the time to complete the task and the number of errors. Those two measures were the most common in the reviewed robotic visual assistant studies. For the \textsc{Reachability} and \textsc{Manipulability} tasks, the number of errors is the number of manipulator collisions. For the \textsc{Passability} task, the number of errors is the number of robot collisions. For the \textsc{Traversability} task, the number of errors is the number of falls of the robot.

The metric indicating the quality of a viewpoint is the subject's performance computed as
\begin{equation}
\label{eq:performance}
\tensor*[_j]{P}{^a_i} = -w_t \left(\widetilde{\tensor*[_j]{t}{^a_i}}\right) - w_e \left(\widetilde{\tensor*[_j]{e}{^a_i}}\right),
\end{equation}
where
\begin{equation}
\label{eq:normalized_time}
\widetilde{\tensor*[_j]{t}{^a_i}} = \frac{\tensor*[_j]{t}{^a_i} - \mean_{a^\prime,i^\prime}{\left(\tensor*[_j]{t}{^{a^\prime}_{i^\prime}}\right)}}{\std_{a^\prime,i^\prime}{\left(\tensor*[_j]{t}{^{a^\prime}_{i^\prime}}\right)}},
\end{equation}
\begin{equation}
\label{eq:normalized_errors}
\widetilde{\tensor*[_j]{e}{^a_i}} = \frac{\tensor*[_j]{e}{^a_i} - \mean_{a^\prime,i^\prime}{\left(\tensor*[_j]{e}{^{a^\prime}_{i^\prime}}\right)}}{\std_{a^\prime,i^\prime}{\left(\tensor*[_j]{e}{^{a^\prime}_{i^\prime}}\right)}},
\end{equation}
$j$ is a subject index, $i$ is a viewpoint index, $a$ is an affordance index, $\tensor*[_j]{P}{^a_i}$ denotes the performance of subject $j$ for affordance $a$ from viewpoint $v_i$,  $\tensor*[_j]{t}{^a_i}$ is the time subject $j$ took to complete the task associated with affordance $a$ from viewpoint $v_i$, $\tensor*[_j]{e}{^a_i}$ is the number of errors subject $j$ made when performing the task associated with affordance $a$ from viewpoint $v_i$, $\widetilde{\tensor*[_j]{t}{^a_i}}$ and $\widetilde{\tensor*[_j]{e}{^a_i}}$ are $\tensor*[_j]{t}{^a_i}$ and $\tensor*[_j]{e}{^a_i}$ normalized across all samples for subject $j$, and $w_t$ and $w_e$ are the weights of the time term and error term of the performance respectively. The formula uses time and errors that are normalized for each individual subject to reduce the effects of the variation in overall performance between individual subjects. The weighted sum in this formula is multiplied by $-1$ so that the performance is more intuitive to interpret. Without this adjustment, the lower performance would be better because less time and fewer errors are better, however, lower performance being better is counterintuitive. The $w_t$ and $w_e$ weights are set to $0.4$ and $0.6$ respectively for this experiment. This weights the errors slightly higher than completion time to penalize completing a task faster at the expense of more errors.

The study results in a set of performance samples where one performance sample, $\tensor*[_j]{P}{^a_i}$, represents a performance of subject $j$ at viewpoint $v_i$ for affordance $a$. A performance sample is rejected as an outlier if the corresponding  $\tensor*[_j]{t}{^a_i}$ value is more than three scaled median absolute deviations away from the $\med_{j^\prime}{\tensor*[_{j^\prime}]{t}{^a_i}}$. An example of an outlier would be a subject getting distracted in the middle of a task (e.g., answering a phone call) causing their time to complete the task to be higher than it would have been.

The value of a viewpoint $v_i$ for affordance $a$ is defined as
\begin{equation}
\label{eq:viewpoint_value}
\abs{v_i^a} = w_m \mean_{j}{\left(\tensor*[_j]{P}{^a_i}\right)} - w_d \std_{j}{\left(\tensor*[_j]{P}{^a_i}\right)},
\end{equation}
where $w_m$ and $w_d$ are the weights of the mean and standard deviation term of the viewpoint value respectively. While the viewpoint value should be primarily indicated by the mean of the corresponding performance samples, the standard deviation term is introduced to also make the viewpoint value inversely proportional to the standard deviation of the corresponding performance samples (higher standard deviation indicates higher unpredictability of the performance). The $w_m$ and $w_d$ weights are set to $0.9$ and $0.1$ respectively for this experiment. This makes the mean term the dominant indicator of the viewpoint value.

\subsection{Learning Manifolds by Clustering Viewpoints}
\label{sec:experimentation_clustering}

Agglomerative hierarchical cluster analysis with average linkages~\cite{michener1957a} is used to generate manifolds. Pairwise dissimilarity is computed using the combination of the orthodromic distance and the difference between the normalized viewpoint values, and used to construct a hierarchical cluster tree using the unweighted pair group method with arithmetic mean linkage. The number of manifolds is determined by maximizing the Calinski-Harabasz criterion~\cite{calinski1974a}. The value of a manifold is computed as a combination of the mean and standard deviation of the values of the member viewpoints. To be able to compare two manifolds in terms of time and errors, a metric comparing the time and errors using the intersection of the subjects in the two manifolds is introduced.

The input is 30 sample points for each affordance $a$, where one sample point is
\begin{equation}
\label{eq:sample_point}
s_i^a = (\theta_i, \varphi_i, \widetilde{\abs{v_i^a}}),
\end{equation}
where $\theta_i$ and $\varphi_i$ are the polar angle and azimuthal angle of $i$-th viewpoint, $v_i$, in the spherical coordinate system and $\widetilde{\abs{v_i^a}} = \frac{\abs{v_i^a} - \mean_{i^\prime=1,...,30}{\abs{v_{i^\prime}^a}}}{\std_{i^\prime=1,...,30}{\abs{v_{i^\prime}^a}}}$ is the normalized value of $i$-th viewpoint, $v_i$, for affordance $a$.

The pairwise dissimilarity between all 30 sample points $s_i^a$, where $i = 1, ..., 30$, for affordance $a$ is computed using the combination of the orthodromic distance of the viewpoints on the hemisphere and the difference between the normalized viewpoint values resulting in 435 dissimilarities
\begin{equation}
d_{{s_i^a}{s_j^a}} = \sqrt{\left(d_{s_i^as_j^a}^{(o)}\right)^2 + \left(d_{s_i^as_j^a}^{(p)}\right)^2},
\end{equation}
where $s_i^a = (\theta_i, \varphi_i, \widetilde{\abs{v_i^a}})$ and $s_j^a = (\theta_j, \varphi_j, \widetilde{\abs{v_j^a}})$ are two sample points, $d_{s_i^as_j^a}^{(o)}$ is the orthodromic distance between the two sample points, and $d_{s_i^as_j^a}^{(p)}$ is the value distance between the two sample points. The orthodromic distance is the great-circle distance of the two associated viewpoints on the hemisphere defined as
\begin{equation}
d_{s_i^as_j^a}^{(o)} = 2r \atantwo{\left(\sqrt{\xi}, \sqrt{1 - \xi}\right)},
\end{equation}
where
\begin{equation}
\xi = \sin^2{\frac{\theta_j - \theta_i}{2}} + \cos{\theta_i} \cos{\theta_j} \sin^2{\frac{\varphi_j - \varphi_i}{2}}.
\end{equation}
The value distance is the difference between the normalized values of the two associated viewpoints defined as
\begin{equation}
d_{s_i^as_j^a}^{(p)} = \abs{\widetilde{\abs{v_i^a}} - \widetilde{\abs{v_j^a}}}.
\end{equation}

The sample points are grouped into a binary hierarchical cluster tree using the unweighted pair group method with arithmetic mean linkage~\cite{michener1957a}. The linkage between clusters (manifolds) $M_k^a$ and $M_l^a$ for affordance $a$ is the average linkage defined as
\begin{equation}
d_{{M_k^a}{M_l^a}} = \frac{1}{N_{M_k^a}N_{M_l^a}} \sum_{s_i^a \in M_k^a} \sum_{s_j^a \in M_l^a} d_{{s_i^a}{s_j^a}},
\end{equation}
where $N_{M_k^a}$ is the number of sample points in $k$-th cluster (manifold), $M_k^a$, for affordance $a$.

The number of manifolds $N_m^a$ for affordance $a$ is determined by maximizing the Calinski-Harabasz criterion~\cite{calinski1974a}. This criterion is used based on a premise that well-defined clusters have a large between-cluster variance and a small within-cluster variance. For this work, the number of manifolds is limited to $10$ to prevent forming too many small manifolds.

The value of a manifold $M_k^a$ is defined as
\begin{equation}
\label{eq:manifold_value}
\abs{M_k^a} = w_m \mean_{v_i \in M_k^a}{\abs{v_i^a}} - w_d \std_{v_i \in M_k^a}{\abs{v_i^a}},
\end{equation}
where $v_i \in M_k^a \iff s_i^a \in M_k^a$.

To be able to compare two manifolds in terms of the non-normalized time and number of errors, a metric quantifying a relative improvement in the time and number of errors between two manifolds is introduced. This metric measures the improvement only on the intersection of the subjects in the two manifolds to prevent biasing the relative improvement with the variation in overall speed among individual subjects and is defined as
\begin{equation}
\label{eq:relative_improvement}
I_{M_k^aM_l^a}^{(t)} = \frac{\tensor*[_{M_k^a}]{t}{^a_{M_l^a}} - \tensor*[_{M_l^a}]{t}{^a_{M_k^a}}}{\tensor*[_{M_k^a}]{t}{^a_{M_l^a}}},
\end{equation}
where $I_{M_k^aM_l^a}^{(t)}$ is the relative improvement in time $t$ of manifold $M_k^a$ over manifold $M_l^a$, $\tensor*[_{M_l^a}]{t}{^a_{M_k^a}} = \mean_{j \in \mathcal{S}\left(M_k^a\right) \cap \mathcal{S}\left(M_l^a\right)}{\tensor*[_j]{t}{^a_{M_k^a}}}$ is the average time to complete the task associated with affordance $a$ from manifold $M_k^a$ measured by only taking subjects that have at least one sample in both $M_k^a$ and $M_l^a$ manifolds, $\tensor*[_j]{t}{^a_{M_k^a}} = \mean_{v_i \in M_k^a}{\tensor*[_j]{t}{^a_i}}$ is the average time subject $j$ took to complete the task associated with affordance $a$ from manifold $M_k^a$, and $\mathcal{S}\left(M_k^a\right)$ is the set of subjects that have at least one sample in manifold $M_k^a$. The relative improvement in the errors $I_{M_k^aM_l^a}^{(e)}$ is measured analogically.

\section{Results}
\label{sec:results}

The results show the validity of the affordance-based approach by confirming there are manifolds of statistically significantly different viewpoint values, viewpoint values depend on the affordances, and viewpoint values are independent of a robot. The best manifold for each affordance provides a statistically significant improvement with a large Cohen's $d$ effect size (1.1--2.3) in performance (improving time by 14\%--59\% and reducing errors by 87\%--100\%) and improvement in performance variation over the worst manifold. All statistical testing is on significance level $\alpha = 0.05$.

\subsection{Validity of Affordance-Based Approach}

The results support the two central tenets of the approach by confirming that there are manifolds of statistically significantly different viewpoint values, viewpoint values are statistically significantly dependent on the affordances, and viewpoint values are independent of a robot.

There are manifolds of statistically significantly different viewpoint values. Not all views are equal, and some manifolds provide statistically significantly better views than others. This is tested using an unbalanced one-way analysis of variance (ANOVA) test for each affordance testing that not all $P_{M_k^a}^a$ for $k = 1, ..., N_m^a$ for a specific affordance $a$ are equal, where $P_{M_k^a}^a = \mean_{v_i \in M_k^a}{\left(\tensor*[_j]{P}{^a_i}\right)}$ is the mean of all performance samples in manifold $M_k^a$ for affordance $a$. This is confirmed for all affordances $a$ based on $F$-statistics and $p$-values listed in Table~\ref{tab:manifolds_different}.

\begin{table*}[!t]
\caption{There Are Manifolds of Statistically Significantly Different Viewpoint Values}
\label{tab:manifolds_different}
\resizebox{\textwidth}{!}{
\renewcommand{\arraystretch}{1.3}
\setlength\tabcolsep{1.5pt}
\begin{tabular}{|l|>{\centering\arraybackslash}m{0.92cm};{.4pt/1pt}>{\centering\arraybackslash}m{0.92cm}|c;{.4pt/1pt}c;{.4pt/1pt}c;{.4pt/1pt}c;{.4pt/1pt}c;{.4pt/1pt}c|c;{.4pt/1pt}c;{.4pt/1pt}c;{.4pt/1pt}c;{.4pt/1pt}c;{.4pt/1pt}c;{.4pt/1pt}c;{.4pt/1pt}c;{.4pt/1pt}c;{.4pt/1pt}c|c;{.4pt/1pt}c;{.4pt/1pt}c;{.4pt/1pt}c;{.4pt/1pt}c;{.4pt/1pt}c;{.4pt/1pt}c|}
\hline
\rowcolor[HTML]{AFABAB} 
\textbf{Affordance} $\left(a\right) \blacktriangleright$ & \multicolumn{2}{c|}{\cellcolor[HTML]{AFABAB}\textbf{\textsc{Reachability}}} & \multicolumn{6}{c|}{\cellcolor[HTML]{AFABAB}\textbf{\textsc{Passability}}} & \multicolumn{10}{c|}{\cellcolor[HTML]{AFABAB}\textbf{\textsc{Manipulability}}} & \multicolumn{7}{c|}{\cellcolor[HTML]{AFABAB}\textbf{\textsc{Traversability}}} \\ \hline
\textbf{Number of Manifolds} $\left(N_m^a\right)$ & \multicolumn{2}{c|}{$2$} & \multicolumn{6}{c|}{$6$} & \multicolumn{10}{c|}{$10$} & \multicolumn{7}{c|}{$7$} \\ \hline
\rowcolor[HTML]{E7E6E6} 
\textbf{Manifold} $\left(M_k^a\right)$ $\blacktriangleright$ & $M_1^\textsc{R}$ & $M_2^\textsc{R}$ & $M_1^\textsc{P}$ & $M_2^\textsc{P}$ & $M_3^\textsc{P}$ & $M_4^\textsc{P}$ & $M_5^\textsc{P}$ & $M_6^\textsc{P}$ & $M_1^\textsc{M}$ & $M_2^\textsc{M}$ & $M_3^\textsc{M}$ & $M_4^\textsc{M}$ & $M_5^\textsc{M}$ & $M_6^\textsc{M}$ & $M_7^\textsc{M}$ & $M_8^\textsc{M}$ & $M_9^\textsc{M}$ & $M_{10}^\textsc{M}$ & $M_1^\textsc{T}$ & $M_2^\textsc{T}$ & $M_3^\textsc{T}$ & $M_4^\textsc{T}$ & $M_5^\textsc{T}$ & $M_6^\textsc{T}$ & $M_7^\textsc{T}$ \\ \hline
\textbf{Number of Samples} $\left(N_{M_k^a}^{(P)}\right)$ & $109$ & $33$ & $34$ & $47$ & $18$ & $5$ & $25$ & $12$ & $5$ & $15$ & $24$ & $18$ & $29$ & $24$ & $14$ & $14$ & $6$ & $1$ & $29$ & $17$ & $28$ & $39$ & $15$ & $14$ & $1$ \\ \hline
\textbf{Performance Mean} $\left(P_{M_k^a}^a\right)$ & $0.52$ & $0.09$ & $0.5$ & $0.46$ & $0.35$ & $-0.01$ & $-0.25$ & $-0.5$  & $0.27$ & $0.09$ & $0.05$ & $0.01$ & $-0.45$ & $-0.55$ & $-0.51$ & $-0.67$ & $-1.5$ & $-1.8$ & $0.26$ & $0.27$ & $-0.13$ & $-0.33$ & $-0.71$ & $-0.97$ & $-2.6$ \\ \hline
\textbf{$F$-statistic} & \multicolumn{2}{c|}{\begin{tabular}[c]{@{}l@{}}$F(1, 140)$\\ $=30.3359$\end{tabular}} & \multicolumn{6}{c|}{$F(5, 135)=19.1088$} & \multicolumn{10}{c|}{$F(9, 140)=5.4869$} & \multicolumn{7}{c|}{$F(6, 136)=7.4848$} \\ \hline
\textbf{$p$-value} & \multicolumn{2}{c|}{$1.6872\times10^{-7}$} & \multicolumn{6}{c|}{$2.4179\times10^{-14}$} & \multicolumn{10}{c|}{$1.8816\times10^{-6}$} & \multicolumn{7}{c|}{$6.083\times10^{-7}$} \\ \hline
\end{tabular}
}
\end{table*}

The viewpoint values are statistically significantly dependent on the affordances. This is tested using an unbalanced two-way ANOVA test for interaction effects testing whether there is an interaction between affordance factor $a$ and viewpoint factor $i$ for response variable $\widetilde{P_{v_i}^a}$, where $\widetilde{P_{v_i}^a} = \mean_{j}{\left(\widetilde{\tensor*[_j]{P}{^a_i}}\right)}$ is the mean normalized performance for viewpoint $v_i$ for affordance $a$ and $\widetilde{\tensor*[_j]{P}{^a_i}} = \frac{\tensor*[_j]{P}{^a_i} - \mean_{i^\prime,j^\prime}{\left(\tensor*[_{j^\prime}]{P}{^a_{i^\prime}}\right)}}{\std_{i^\prime,j^\prime}{\left(\tensor*[_{j^\prime}]{P}{^a_{i^\prime}}\right)}}$ is the performance $\tensor*[_j]{P}{^a_i}$ normalized within the affordance $a$ (normalization is necessary because different affordances have different scales of performance). The interaction is confirmed based on $F$-statistic $F(86, 457) = 1.8361$ and $p$-value $4.0652 \times 10^{-5}$.

The viewpoint values are independent of the robot. The selected robot does not have a statistically significant influence on viewpoint values (i.e., the viewpoint values are very similar for both robots). This is tested using an unbalanced two-way ANOVA test for interaction effects testing whether there is an interaction between robot factor $\rho$ and viewpoint factor $i$ for response variable $\widetilde{\tensor*[^\rho]{P}{_{v_i}}}$, where $\widetilde{\tensor*[^\rho]{P}{_{v_i}}} = \mean_{j,a}{\left(\widetilde{\tensor*[_j^\rho]{P}{^a_i}}\right)}$ is the mean normalized performance for viewpoint $v_i$ for robot $\rho$, $\widetilde{\tensor*[_j^\rho]{P}{^a_i}} = \frac{\tensor*[_j^\rho]{P}{^a_i} - \mean_{i^\prime,j^\prime,\rho^\prime}{\left(\tensor*[_{j^\prime}^{\rho^\prime}]{P}{^a_{i^\prime}}\right)}}{\std_{i^\prime,j^\prime,\rho^\prime}{\left(\tensor*[_{j^\prime}^{\rho^\prime}]{P}{^a_{i^\prime}}\right)}}$ is the $\tensor*[_j^\rho]{P}{^a_i}$ normalized within the affordance $a$ (normalization is necessary because different affordances have different scales of performance), and $\tensor*[_j^\rho]{P}{^a_i}$ is the performance of subject $j$ for affordance $a$ and robot $\rho$ from viewpoint $v_i$. The interaction is not confirmed based on $F$-statistic $F(29, 516) = 1.1507$ and $p$-value $0.27078$. Since the interaction term of the ANOVA test is non-significant, the interaction effect is either very small and statistically non-significant or does not exist.

\subsection{Statistically Significant Improvement in Performance}

The results show there is a statistically significant improvement with a large Cohen's $d$ effect size (1.1--2.3) between the best and worst manifold for each affordance improving time by 14\% to 59\% and reducing errors by 87\% to 100\%. The best manifolds also provide an improvement in performance variation over the worst manifolds.

\begin{figure}[!t]
\centering
\includegraphics[width=3in]{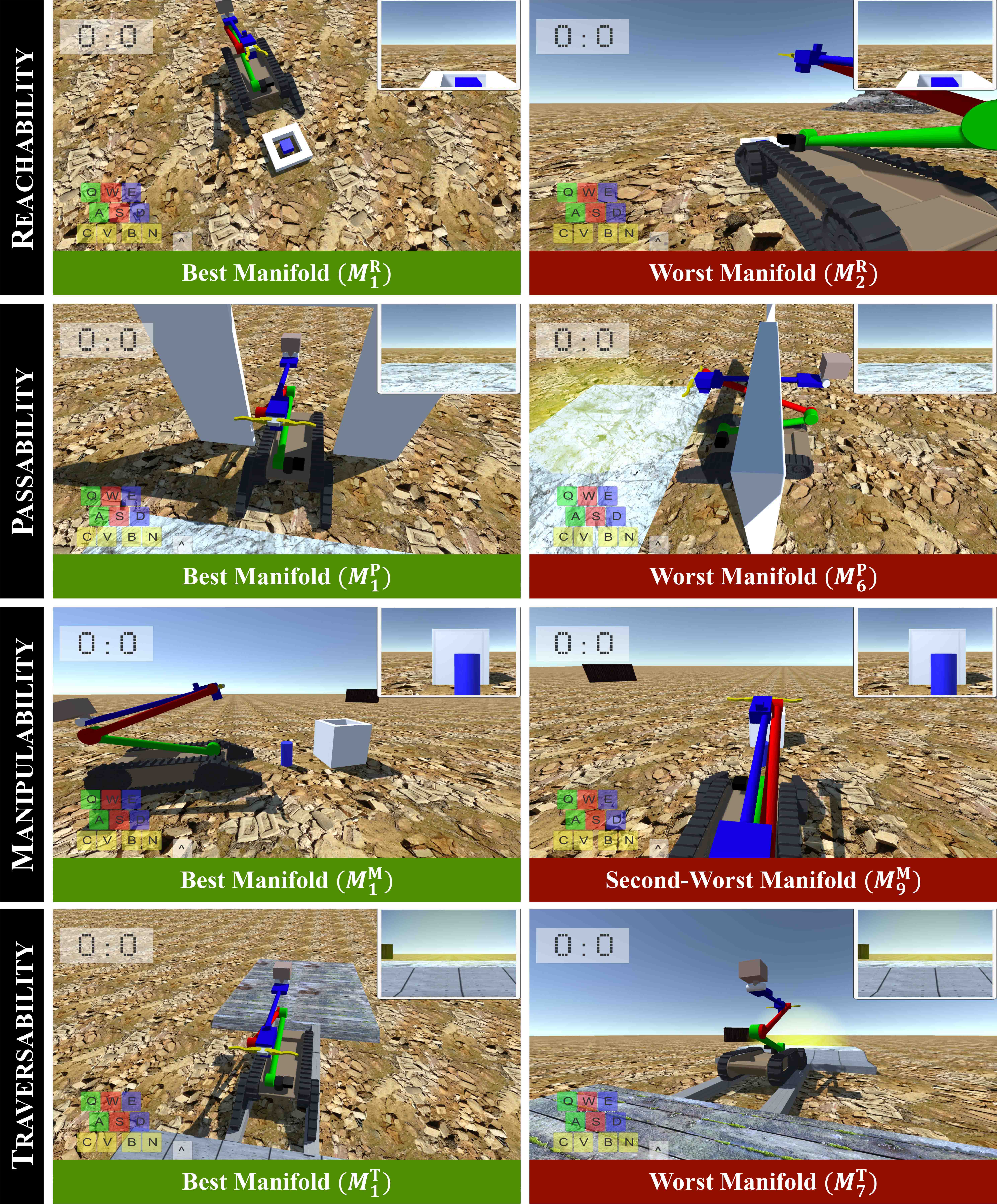}
\caption{The view from the centroid of the best manifold as compared to the worst manifold.}
\label{fig:view_best_worst}
\end{figure}

\begin{figure}[!t]
\centering
\includegraphics[width=2.5in]{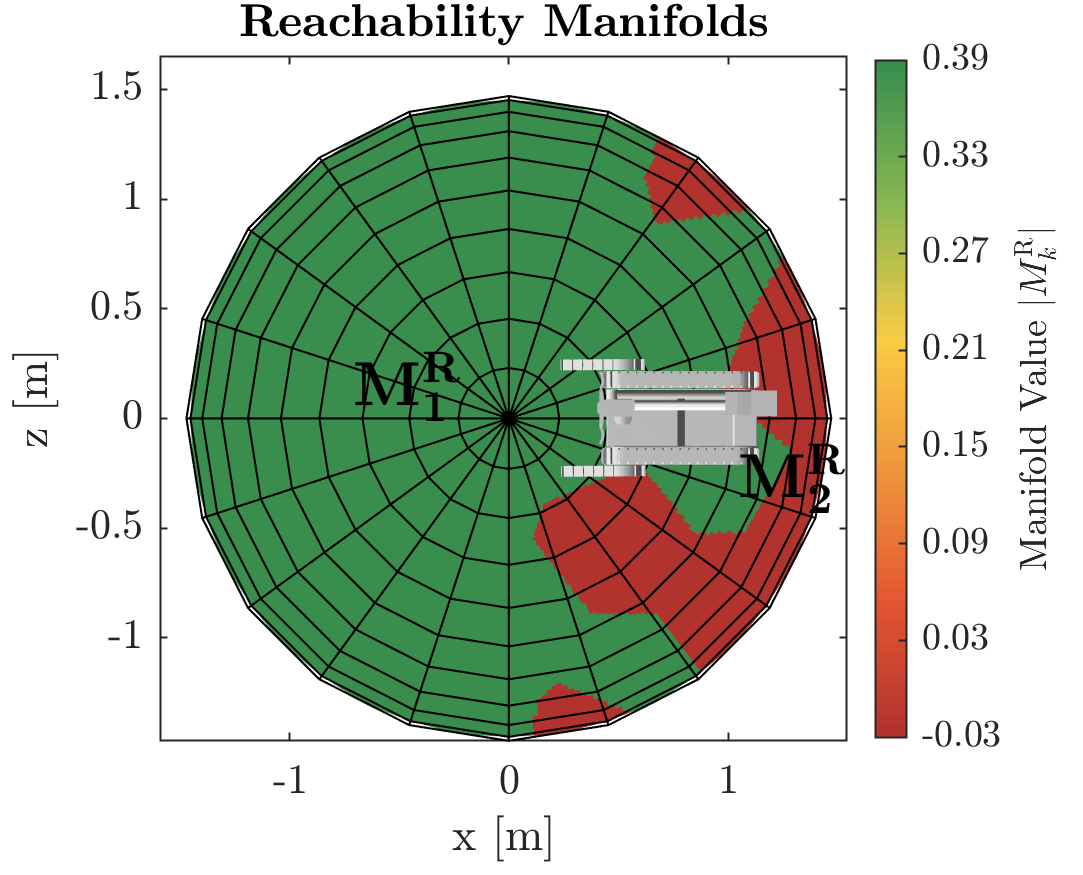}
\caption{The 2 manifolds for \textsc{Reachability} affordance. The figure shows a top-down view of the hemisphere with the 2 manifolds ($M_1^\textsc{R}$ and $M_2^\textsc{R}$) color-coded by the corresponding manifold value ($\abs{M_1^\textsc{R}}$ and $\abs{M_2^\textsc{R}}$). The manifold labels are placed in the centroid of the corresponding manifold. The manifolds are indexed in ascending order from the best to the worst. The figure is in scale including the size and pose of the primary robot (colored gray).}
\label{fig:reachability_manifolds}
\end{figure}

\begin{figure}[!t]
\centering
\includegraphics[width=2.5in]{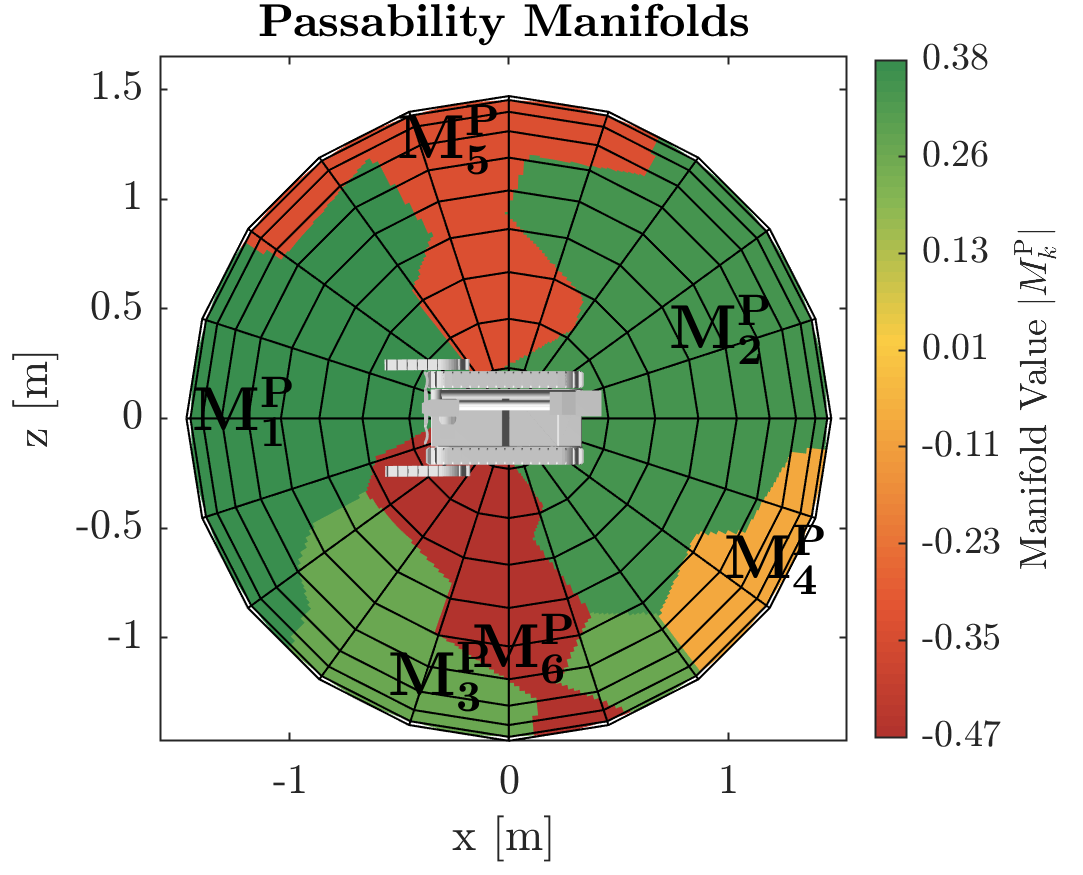}
\caption{The 6 manifolds for \textsc{Passability} affordance. The figure shows a top-down view of the hemisphere with the 6 manifolds ($M_1^\textsc{P}, ..., M_6^\textsc{P}$) color-coded by the corresponding manifold value ($\abs{M_1^\textsc{P}}, ..., \abs{M_6^\textsc{P}}$).}
\label{fig:passability_manifolds}
\end{figure}

\begin{figure}[!t]
\centering
\includegraphics[width=2.5in]{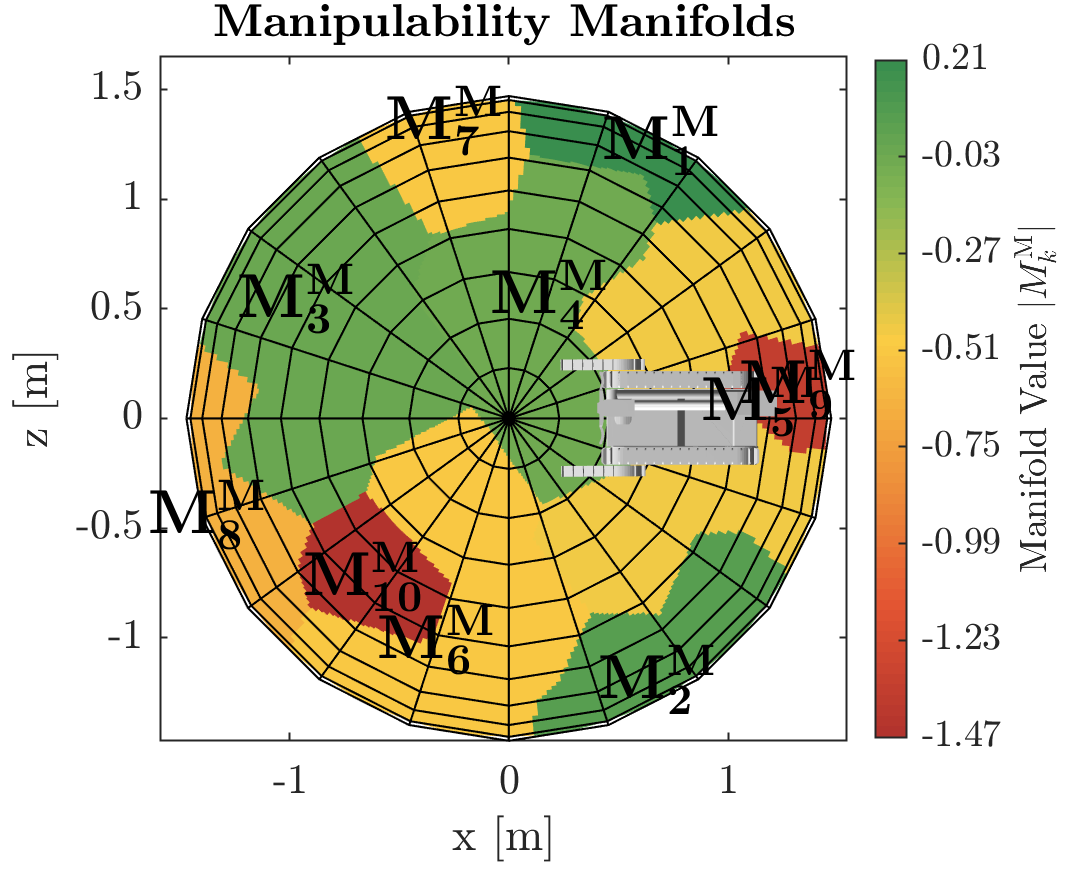}
\caption{The 10 manifolds for \textsc{Manipulability} affordance. The figure shows a top-down view of the hemisphere with the 10 manifolds ($M_1^\textsc{M}, ..., M_{10}^\textsc{M}$) color-coded by the corresponding manifold value ($\abs{M_1^\textsc{M}}, ..., \abs{M_{10}^\textsc{M}}$).}
\label{fig:manipulability_manifolds}
\end{figure}

\begin{figure}[!t]
\centering
\includegraphics[width=2.5in]{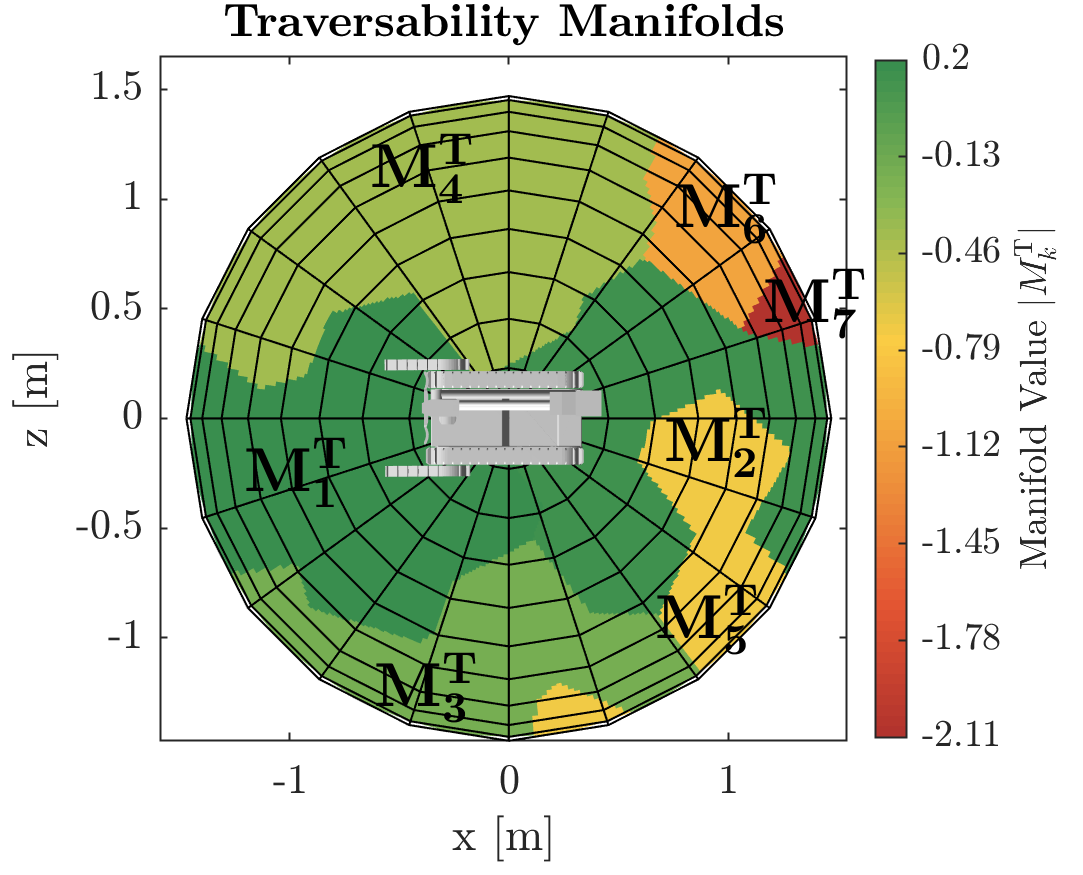}
\caption{The 7 manifolds for \textsc{Traversability} affordance. The figure shows a top-down view of the hemisphere with the 7 manifolds ($M_1^\textsc{T}, ..., M_7^\textsc{T}$) color-coded by the corresponding manifold value ($\abs{M_1^\textsc{T}}, ..., \abs{M_7^\textsc{T}}$).}
\label{fig:traversability_manifolds}
\end{figure}

The best manifold for each affordance improves time by 14\% to 59\% and reduces errors by 87\% to 100\% over the worst manifold. Figure~\ref{fig:view_best_worst} shows for each affordance a view from the best manifold as compared to the worst manifold. Tables~\ref{tab:passability_results}--\ref{tab:traversability_results} provide a quantitative comparison between the best and worst manifold for each affordance (for \textsc{Manipulability}, the second-worst manifold is used for this comparison as there are no subjects in the intersection of the worst and best manifold). Figures~\ref{fig:reachability_manifolds}--\ref{fig:traversability_manifolds} show the visualization of the manifolds for each affordance.

\begin{table}[!t]
\caption{\textsc{Reachability}: 14\% Time and 87\% Error Improvement}
\label{tab:reachability_results}
\resizebox{0.49\textwidth}{!}{
\renewcommand{\arraystretch}{1.3}
\begin{tabular}{|l|c;{.4pt/1pt}c|c|}
\hline
\rowcolor[HTML]{AFABAB} 
\textbf{Affordance} $\left(a\right)$ $\blacktriangleright$                             & \multicolumn{3}{c|}{\cellcolor[HTML]{AFABAB}\textbf{\textsc{Reachability}}}                                                                                                                                                                                                       \\ \hline
\textbf{Number of Manifolds} $\left(N_m^a\right)$ & \multicolumn{3}{c|}{$2$}                                                                                                                                                                                                                                                   \\ \hline
\rowcolor[HTML]{E7E6E6}
\textbf{Manifold} $\left(M_k^a\right)$ $\blacktriangleright$         & \textbf{Best} $\left(M_1^\textsc{R}\right)$                                                                & \textbf{Worst} $\left(M_2^\textsc{R}\right)$                                                               & \textbf{Improvement} $\left(I_{M_1^\textsc{R}M_2^\textsc{R}}\right)$ \\ \hline
\textbf{Manifold Value} $\left(\abs{M_k^a}\right)$   & $0.387$                                                                                                  & $-0.032$                                                                                                 & -                                                      \\ \hline
\rowcolor[HTML]{DBE2F1} 
\textbf{Time}                                          & $\tensor*[_{M_2^\textsc{R}}]{t}{^{\textsc{R}}_{M_1^\textsc{R}}} = \SI{21.11}{\second}$ & $\tensor*[_{M_1^\textsc{R}}]{t}{^{\textsc{R}}_{M_2^\textsc{R}}} = \SI{24.58}{\second}$ & \cellcolor[HTML]{4E6FBB}{\color[HTML]{FFFFFF} \textbf{14\%}}                  \\ \hline
\rowcolor[HTML]{DBD3E8} 
\textbf{Errors}                                        & $\tensor*[_{M_2^\textsc{R}}]{e}{^{\textsc{R}}_{M_1^\textsc{R}}} = 0.08$   & $\tensor*[_{M_1^\textsc{R}}]{e}{^{\textsc{R}}_{M_2^\textsc{R}}} = 0.613$   & \cellcolor[HTML]{4A1D8B}{\color[HTML]{FFFFFF} \textbf{87\%}}                  \\ \hline
\end{tabular}
}
\end{table}

\begin{table}[!t]
\caption{\textsc{Passability}: 23\% Time and 100\% Error Improvement}
\label{tab:passability_results}
\resizebox{0.49\textwidth}{!}{
\renewcommand{\arraystretch}{1.3}
\begin{tabular}{|l|c;{.4pt/1pt}c|c|}
\hline
\rowcolor[HTML]{AFABAB} 
\textbf{Affordance} $\left(a\right)$ $\blacktriangleright$                             & \multicolumn{3}{c|}{\cellcolor[HTML]{AFABAB}\textbf{\textsc{Passability}}}                                                                                                                                                                                                       \\ \hline
\textbf{Number of Manifolds} $\left(N_m^a\right)$ & \multicolumn{3}{c|}{$6$}                                                                                                                                                                                                                                                   \\ \hline
\rowcolor[HTML]{E7E6E6}
\textbf{Manifold} $\left(M_k^a\right)$ $\blacktriangleright$         & \textbf{Best} $\left(M_1^\textsc{P}\right)$                                                                & \textbf{Worst} $\left(M_6^\textsc{P}\right)$                                                               & \textbf{Improvement} $\left(I_{M_1^\textsc{P}M_6^\textsc{P}}\right)$ \\ \hline
\textbf{Manifold Value} $\left(\abs{M_k^a}\right)$   & $0.376$                                                                                                  & $-0.47$                                                                                                 & -                                                      \\ \hline
\rowcolor[HTML]{DBE2F1} 
\textbf{Time}                                          & $\tensor*[_{M_6^\textsc{P}}]{t}{^{\textsc{P}}_{M_1^\textsc{P}}} = \SI{20.53}{\second}$ & $\tensor*[_{M_1^\textsc{P}}]{t}{^{\textsc{P}}_{M_6^\textsc{P}}} = \SI{26.74}{\second}$ & \cellcolor[HTML]{4E6FBB}{\color[HTML]{FFFFFF} \textbf{23\%}}                  \\ \hline
\rowcolor[HTML]{DBD3E8} 
\textbf{Errors}                                        & $\tensor*[_{M_6^\textsc{P}}]{e}{^{\textsc{P}}_{M_1^\textsc{P}}} = 0$   & $\tensor*[_{M_1^\textsc{P}}]{e}{^{\textsc{P}}_{M_6^\textsc{P}}} = 1.2$   & \cellcolor[HTML]{4A1D8B}{\color[HTML]{FFFFFF} \textbf{100\%}}                  \\ \hline
\end{tabular}
}
\end{table}

\begin{table}[!t]
\caption{\textsc{Manipulability}: 39\% Time and 100\% Error Improvement}
\label{tab:manipulability_results}
\resizebox{0.49\textwidth}{!}{
\renewcommand{\arraystretch}{1.3}
\begin{tabular}{|l|c;{.4pt/1pt}c|c|}
\hline
\rowcolor[HTML]{AFABAB} 
\textbf{Affordance} $\left(a\right)$ $\blacktriangleright$                             & \multicolumn{3}{c|}{\cellcolor[HTML]{AFABAB}\textbf{\textsc{Manipulability}}}                                                                                                                                                                                                       \\ \hline
\textbf{Number of Manifolds} $\left(N_m^a\right)$ & \multicolumn{3}{c|}{$10$}                                                                                                                                                                                                                                                   \\ \hline
\rowcolor[HTML]{E7E6E6}
\textbf{Manifold} $\left(M_k^a\right)$ $\blacktriangleright$         & \textbf{Best} $\left(M_1^\textsc{M}\right)$                                                                & \textbf{Second-Worst} $\left(M_{9}^\textsc{M}\right)$                                                               & \textbf{Improvement} $\left(I_{M_1^\textsc{M}M_{9}^\textsc{M}}\right)$ \\ \hline
\textbf{Manifold Value} $\left(\abs{M_k^a}\right)$   & $0.212$                                                                                                  & $-1.379$                                                                                                 & -                                                      \\ \hline
\rowcolor[HTML]{DBE2F1} 
\textbf{Time}                                          & $\tensor*[_{M_{9}^\textsc{M}}]{t}{^{\textsc{M}}_{M_1^\textsc{M}}} = \SI{29.57}{\second}$ & $\tensor*[_{M_1^\textsc{M}}]{t}{^{\textsc{M}}_{M_{9}^\textsc{M}}} = \SI{48.84}{\second}$ & \cellcolor[HTML]{4E6FBB}{\color[HTML]{FFFFFF} \textbf{39\%}}                  \\ \hline
\rowcolor[HTML]{DBD3E8} 
\textbf{Errors}                                        & $\tensor*[_{M_{9}^\textsc{M}}]{e}{^{\textsc{M}}_{M_1^\textsc{M}}} = 0$   & $\tensor*[_{M_1^\textsc{M}}]{e}{^{\textsc{M}}_{M_{9}^\textsc{M}}} = 0.5$   & \cellcolor[HTML]{4A1D8B}{\color[HTML]{FFFFFF} \textbf{100\%}}                  \\ \hline
\end{tabular}
}
\end{table}

\begin{table}[!t]
\caption{\textsc{Traversability}: 59\% Time and 100\% Error Improvement}
\label{tab:traversability_results}
\resizebox{0.49\textwidth}{!}{
\renewcommand{\arraystretch}{1.3}
\begin{tabular}{|l|c;{.4pt/1pt}c|c|}
\hline
\rowcolor[HTML]{AFABAB} 
\textbf{Affordance} $\left(a\right)$ $\blacktriangleright$                             & \multicolumn{3}{c|}{\cellcolor[HTML]{AFABAB}\textbf{\textsc{Traversability}}}                                                                                                                                                                                                       \\ \hline
\textbf{Number of Manifolds} $\left(N_m^a\right)$ & \multicolumn{3}{c|}{$7$}                                                                                                                                                                                                                                                   \\ \hline
\rowcolor[HTML]{E7E6E6}
\textbf{Manifold} $\left(M_k^a\right)$ $\blacktriangleright$         & \textbf{Best} $\left(M_1^\textsc{T}\right)$                                                                & \textbf{Worst} $\left(M_7^\textsc{T}\right)$                                                               & \textbf{Improvement} $\left(I_{M_1^\textsc{T}M_7^\textsc{T}}\right)$ \\ \hline
\textbf{Manifold Value} $\left(\abs{M_k^a}\right)$   & $0.197$                                                                                                  & $-2.108$                                                                                                 & -                                                      \\ \hline
\rowcolor[HTML]{DBE2F1} 
\textbf{Time}                                          & $\tensor*[_{M_7^\textsc{T}}]{t}{^{\textsc{T}}_{M_1^\textsc{T}}} = \SI{23.47}{\second}$ & $\tensor*[_{M_1^\textsc{T}}]{t}{^{\textsc{T}}_{M_7^\textsc{T}}} = \SI{57.02}{\second}$ & \cellcolor[HTML]{4E6FBB}{\color[HTML]{FFFFFF} \textbf{59\%}}                  \\ \hline
\rowcolor[HTML]{DBD3E8} 
\textbf{Errors}                                        & $\tensor*[_{M_7^\textsc{T}}]{e}{^{\textsc{T}}_{M_1^\textsc{T}}} = 0$   & $\tensor*[_{M_1^\textsc{T}}]{e}{^{\textsc{T}}_{M_7^\textsc{T}}} = 5$   & \cellcolor[HTML]{4A1D8B}{\color[HTML]{FFFFFF} \textbf{100\%}}                  \\ \hline
\end{tabular}
}
\end{table}

The best manifold is statistically significantly better than the worst manifold for each affordance with a large Cohen's $d$ effect size (1.1--2.3). This is tested using one-tailed two-sample t-test (left-tailed) testing for each affordance $a$ that $P_{M_{w(a)}^a}^a < P_{M_{b(a)}^a}^a$, where $M_{w(a)}^a$ and $M_{b(a)}^a$ are the worst and best manifolds for affordance $a$ respectively, $w(a) = \argmin_{k = 1,...,N_m^a}{\abs{M_k^a}}$, and $b(a) = \argmax_{k = 1,...,N_m^a}{\abs{M_k^a}}$. The worst manifold for \textsc{Manipulability} ($M_{10}^\textsc{M}$) and \textsc{Traversability} ($M_7^\textsc{T}$) was replaced by the second-worst manifold, $M_9^\textsc{M}$ and $M_6^\textsc{T}$ respectively, because the worst manifolds each have only one sample. The hypothesis is confirmed for all affordances $a$ based on $t$-statistics and $p$-values listed in Table~\ref{tab:best_manifold_better_than_worst_manifold}. The table also lists Cohen's $d$ effect size, number of performance samples $N_{M_k^a}^{(P)}$, and the standard deviation of performance $\sigma\left(P_{M_k^a}^a\right) = \std_{v_i \in M_k^a}{\left(\tensor*[_j]{P}{^a_i}\right)}$ for manifold $M_k^a$.

\begin{table*}[!t]
\caption{Best Manifolds Statistically Significantly Better Than Worst with Large Cohen's $d$ Effect Size}
\label{tab:best_manifold_better_than_worst_manifold}
\centering
\resizebox{0.9\textwidth}{!}{
\renewcommand{\arraystretch}{1.2}
\setlength\tabcolsep{2pt}
\begin{tabular}{|l|c;{.4pt/1pt}c|c;{.4pt/1pt}c|c;{.4pt/1pt}c|c;{.4pt/1pt}c|}
\hline
\rowcolor[HTML]{AFABAB} 
\textbf{Affordance} $\left(a\right)$ $\blacktriangleright$                                                                    & \multicolumn{2}{c|}{\cellcolor[HTML]{AFABAB}\textbf{\textsc{Reachability}}}                  & \multicolumn{2}{c|}{\cellcolor[HTML]{AFABAB}\textbf{\textsc{Passability}}}                   & \multicolumn{2}{c|}{\cellcolor[HTML]{AFABAB}\textbf{\textsc{Manipulability}}}                & \multicolumn{2}{c|}{\cellcolor[HTML]{AFABAB}\textbf{\textsc{Traversability}}}                 \\ \hline
\textbf{Number of Manifolds} $\left(N_m^a\right)$                                        & \multicolumn{2}{c|}{$2$}                                                              & \multicolumn{2}{c|}{$6$}                                                              & \multicolumn{2}{c|}{$10$}                                                             & \multicolumn{2}{c|}{$7$}                                                               \\ \hline
\rowcolor[HTML]{E7E6E6} 
\textbf{Manifold} $\left(M_k^a\right)$ $\blacktriangleright$                                                & \textbf{Best} $\left(M_1^\textsc{R}\right)$ & \textbf{Worst} $\left(M_2^\textsc{R}\right)$ & \textbf{Best} $\left(M_1^\textsc{P}\right)$ & \textbf{Worst} $\left(M_6^\textsc{P}\right)$ & \textbf{Best} $\left(M_1^\textsc{M}\right)$ & \textbf{Worst} $\left(M_9^\textsc{M}\right)^*$ & \textbf{Best} $\left(M_1^\textsc{T}\right)$ & \textbf{Worst} $\left(M_6^\textsc{T}\right)^*$ \\ \hline
\textbf{Number of Samples} $\left(N_{M_k^a}^{(P)}\right)$             & $109$                                      & $33$                                       & $34$                                       & $12$                                       & $5$                                       & $6$                                         & $29$                                       & $14$                                        \\ \hline
\textbf{Performance Mean} $\left(P_{M_k^a}^a\right)$                  & $0.51582$                                  & $0.093026$                                 & $0.50487$                                  & $-0.49656$                                 & $0.27369$                                 & $-1.5455$                                   & $0.25833$                                  & $-0.96995$                                  \\ \hline
\textbf{Performance Std. Dev.} $\left(\sigma\left(P_{M_k^a}^a\right)\right)$ & $0.25781$                                  & $0.65474$                                  & $0.13714$                                  & $0.84357$                                  & $0.097376$                                & $1.4138$                                    & $0.2644$                                   & $1.2145$                                    \\ \hline
\textbf{$t$-statistic}                                                                          & \multicolumn{2}{c|}{$-3.6254$}                                                        & \multicolumn{2}{c|}{$-4.0933$}                                                        & \multicolumn{2}{c|}{$-3.1429$}                                                        & \multicolumn{2}{c|}{$-3.7416$}                                                         \\ \hline
\textbf{$p$-value}                                                                              & \multicolumn{2}{c|}{$0.00045367$}                                                     & \multicolumn{2}{c|}{$0.0008565$}                                                      & \multicolumn{2}{c|}{$0.012595$}                                                       & \multicolumn{2}{c|}{$0.0011469$}                                                     \\ \hline
\textbf{Cohen's $d$ Effect Size} $\left(\mathcal{D}^a\right)$                                        & \multicolumn{2}{c|}{$1.0943$}                                                         & \multicolumn{2}{c|}{$2.2854$}                                                         & \multicolumn{2}{c|}{$1.7231$}                                                         & \multicolumn{2}{c|}{$1.7109$}                                                          \\ \hline
\end{tabular}
}
\end{table*}

\begin{figure}[!t]
\begin{center}
\includegraphics[width=0.49\textwidth]{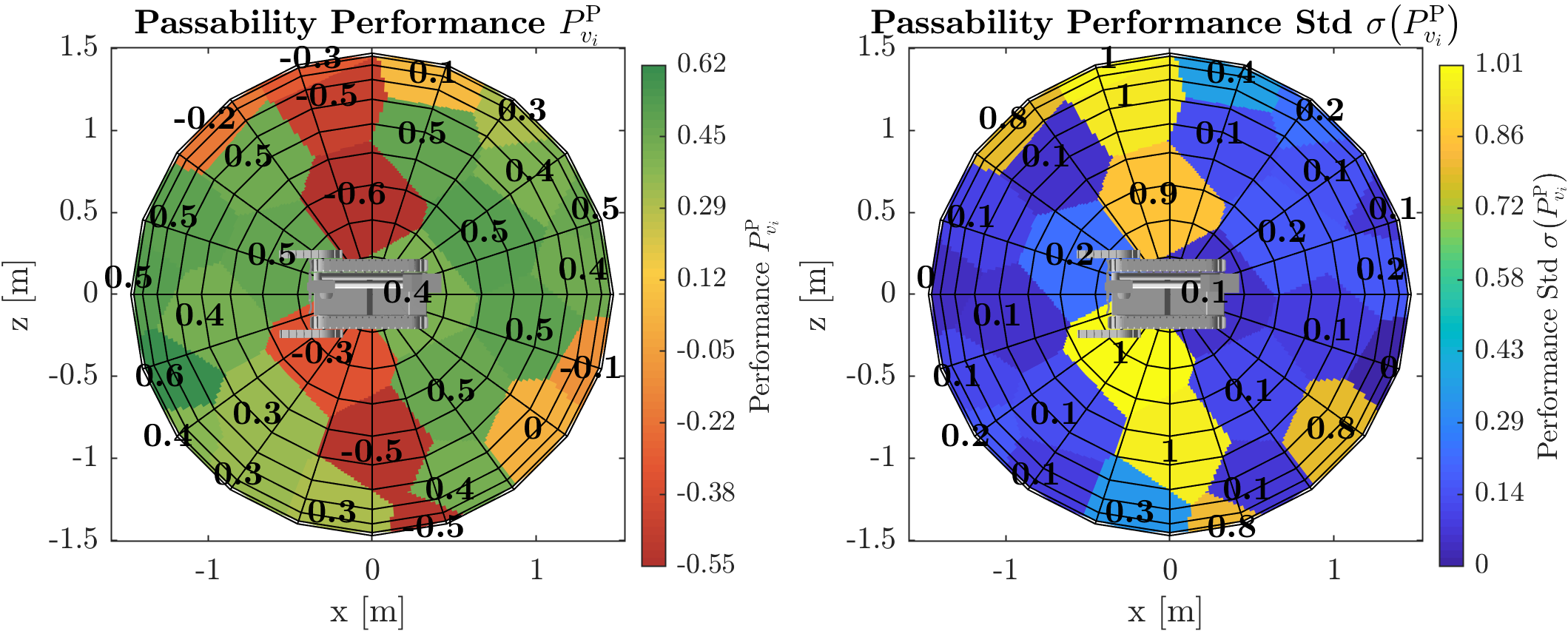}
\end{center}
\caption{The viewpoints with a low mean performance (red) also have a high standard deviation of performance (yellow). The performance (left) and the standard deviation of the performance (right) is defined as $P_{v_i}^a = \mean_j{\left(\tensor*[_j]{P}{^a_i}\right)}$ and $\sigma\left(P_{v_i}^a\right) = \std_j{\left(\tensor*[_j]{P}{^a_i}\right)}$ respectively.}
\label{fig:low_performance_high_standard_deviation}
\end{figure}

The best manifolds provide an improvement in performance variation over the worst manifolds. This is because good viewpoints are consistently good across subjects but bad viewpoints have a large variation in performance (time and errors). This means that having a view from a good manifold leads to good predictable performance while having a view from a bad manifold not only leads to a bad performance but also leads to unpredictability in what might go wrong and how much. Figure~\ref{fig:low_performance_high_standard_deviation} illustrates this on \textsc{Passability} affordance.

\section{Discussion}
\label{sec:discussion}

The results are consistent with Woods et al.\ confirming that an external viewpoint improves teleoperation, and further showing there are manifolds of viewpoints with some manifolds being significantly better than others. While beyond the scope, it is expected the proposed model will likely reduce the cognitive workload by eliminating the need to manually control the robotic visual assistant and improving the ability to comprehend affordances. The results have implications for robotic visual assistants using the manifolds both in terms of visual stability and tracking and can be extracted into actionable rules. The manifolds for all affordances except \textsc{Manipulability} are not very sensitive to the weights used in the viewpoint value and manifold value definitions, it is therefore not necessary to adjust them.

\subsection{Relation to Related Work}

The results are consistent with Woods et al., who showed an external view improves the ability to comprehend affordances making teleoperation easier. This work goes further by showing that not all external views are equal and that there are different regions of viewpoints (manifolds) where good manifolds significantly improve time and reduce errors over bad manifolds. This work is also in agreement with the position of Woods et al.\ that there is no best viewpoint for a task. The viewpoint values in the proposed model depend on the affordances, and therefore the viewpoint would change over time as a task may consist of a series of affordances. For example, a task "go through a door" may consist of four affordances --- \textsc{Traversability} to get to the door, \textsc{Reachability} to reach the door handle, \textsc{Manipulability} to open the door, and \textsc{Passability} to pass through the door (illustrated in Figure~\ref{fig:approach}). In this example, the best viewpoint would change four times during a single task.

\subsection{Reduction in Cognitive Workload}

While beyond the scope of this article, it is expected the model will likely reduce the cognitive workload on the primary robot operator. The model will enable to make the robotic visual assistant autonomous eliminating the need for the primary robot operator to manually control the robotic visual assistant or to coordinate with a secondary operator~\cite{xiao2017visual,xiao2018indoor,xiao2019benchmarking,xiao2018motion,xiao2019explicit,xiao2019robot,xiao2019autonomous,xiao2019explicit2,xiao2020tethered,xiao2020robot}. The model will also allow the robotic visual assistant to select a viewpoint for each action that enables direct apprehension of the affordance for that action reducing the need for high-workload deliberative reasoning about the properties of the scene that would be required if the affordance for the action could not be directly perceived, as shown by Morison~\cite{morison2015human}.

\subsection{Ramifications for Robotic Visual Assistants}

The results have implications for robotic visual assistants using the manifolds both in terms of visual stability and tracking. The best manifolds for all the affordances except \textsc{Manipulability} have large areas (77\%, 23\%, 7\%, and 20\% of the hemisphere surface for \textsc{Reachability}, \textsc{Passability}, \textsc{Manipulability}, \textsc{Traversability} respectively) suggesting that positioning a robotic visual assistant in the best manifold centroid will result in good visual stability (potential perturbations in the pose of a robotic visual assistant will not significantly change the view quality). The best manifolds for \textsc{Reachability} ($M_1^\textsc{R}$) and \textsc{Manipulability} ($M_1^\textsc{M}$ and $M_2^\textsc{M}$) are shifted towards the robot indicating those affordances are object-robot centric and suggesting the necessity to track both the object and the end effector as can be seen in Figures~\ref{fig:reachability_manifolds} and~\ref{fig:manipulability_manifolds} respectively. The best manifolds for \textsc{Passability} ($M_1^\textsc{P}$ and $M_2^\textsc{P}$) and \textsc{Traversability} ($M_1^\textsc{T}$ and $M_2^\textsc{T}$) are elongated along approach and departure directions indicating those affordances are robot-centric and suggesting the necessity to track the entire action (movement) of the robot as can be seen in Figures~\ref{fig:passability_manifolds} and~\ref{fig:traversability_manifolds} respectively. The results show that even small ground-based robotic visual assistants can still provide views from the best manifolds for each affordance since all the manifolds whose value is in the 80th percentile of the manifold value range for the given affordance reach all the way to the ground except for \textsc{Manipulability} (for which the three best manifolds reach the ground but the fourth-best does not).

\subsection{Actionable Rules for Robotic Visual Assistants}

\begin{figure}[!t]
\centering
\includegraphics[width=0.45\textwidth]{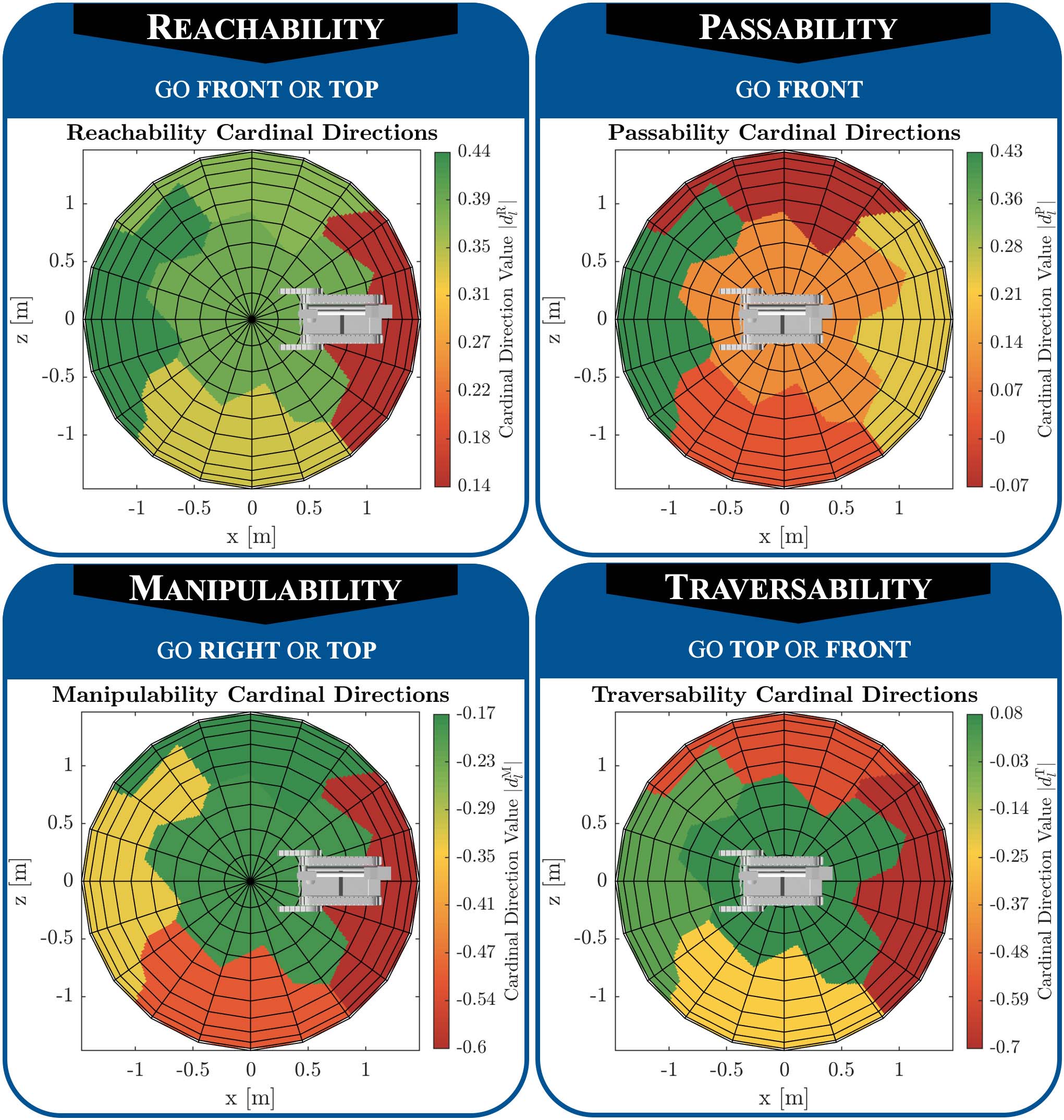}
\caption{Actionable rules for robotic visual assistants and the value for each of the 5 cardinal directions defined as $\abs{d_l^a} = \mean_{v_i \in d_l}{\abs{v_i^a}}$, where $d_l$ denotes $l$-th cardinal direction.}
\label{fig:actionable_rules}
\end{figure}

The results can be extracted into actionable rules for robotic visual assistants suitable for a human operator to follow as shown in Figure~\ref{fig:actionable_rules}. Those rules are extracted by partitioning the viewpoints on the hemisphere into 5 cardinal directions corresponding to the viewpoint groups from Figure~\ref{fig:viewpoints} and computing the view value for each of those cardinal directions as the mean value of the member viewpoints. The desired view direction for each affordance is extracted by taking the cardinal directions whose value is in the 80th percentile of the cardinal direction value range for the given affordance.

\subsection{Sensitivity of Results to Different Weights}

The manifolds for all affordances except \textsc{Manipulability} are not very sensitive to $w_m$ and $w_d$ weights used in the viewpoint value (Equation~\ref{eq:viewpoint_value}) and manifold value (Equation~\ref{eq:manifold_value}) definitions. This is tested by starting with $w_m = 1$ and decreasing $w_m$ by $0.1$ in each step until $w_m = 0.5$ (the other weight is always $w_d = 1 - w_m$). It does not make sense to decrease the $w_m$ weight below $0.5$ as the mean should be the dominant indicator of the viewpoint or manifold value. For \textsc{Reachability} the shape and relative ranking of the manifolds remain unchanged. For \textsc{Passability}, the shape and relative ranking of the manifolds remain unchanged until $w_m = 0.6$. For $w_m = 0.6$ and $w_m = 0.5$, $M_5^\textsc{P}$ has one less viewpoint, otherwise, the shape and relative ranking is unchanged. For \textsc{Traversability}, the shape and relative ranking of the manifolds remain unchanged until $w_m = 0.7$. Then the manifolds start changing, however, the two best manifolds, $M_1^\textsc{T}$ and $M_2^\textsc{T}$, remain unchanged until $w_m = 0.5$ and the best manifold, $M_1^\textsc{T}$, remains unchanged even for $w_m = 0.5$. \textsc{Manipulability} is the most sensitive to different weights, however, the first three manifolds, $M_1^\textsc{M}$, $M_2^\textsc{M}$, and $M_3^\textsc{M}$, remain unchanged in terms of their shape and relative order.

\section{Summary}
\label{sec:summary}

This work proposed a model of the value of different external viewpoints of a robot performing tasks. The model was developed using a psychomotor approach by quantifying the value of 30 external viewpoints for 4 affordances in a study with 31 expert robot operators using a computer-based simulator of two robots and clustering viewpoints of similar value into manifolds of viewpoints with equivalent value using agglomerative hierarchical clustering.

The results support the main postulation of the approach confirming the validity of the affordance-based approach by showing that there are manifolds of statistically significantly different viewpoint values, viewpoint values are statistically significantly dependent on the affordances, and viewpoint values are independent of the robot. The best manifold for each affordance provides a statistically significant improvement with a large Cohen's $d$ effect size (1.1--2.3) in performance and improvement in performance variation over the worst manifold improving time by 14\% to 59\% and reducing errors by 87\% to 100\%.

This work creates the fundamental understanding of external viewpoints quality for four common affordances providing a foundation for ideal viewpoint selection; it is expected, but beyond the scope of this study, that the application of the model to robotic visual assistants will likely reduce the cognitive workload on the primary operator. The model will enable autonomous selection of the best possible viewpoint and path planning for autonomous robotic visual assistants. One direction of future work would be to quantify view quality in a continuous matter rather than for a set of discrete viewpoints.

\section*{Acknowledgment}

Preliminary results were published in~\cite{xiao2019autonomous,xiao2020tethered}. The authors would like to thank Dr. S. Camille Peres for providing valuable feedback and suggestions, and Cassandra Oduola and Mohamed Suhail for their significant work on the simulator and AWS infrastructure introduced in Section~\ref{sec:implementation}.

\ifCLASSOPTIONcaptionsoff
  \newpage
\fi



\bibliographystyle{IEEEtran}
\bibliography{IEEEabrv,references}
\end{document}